\documentclass[journal,twoside,web]{ieeecolor}
\usepackage{generic}
\usepackage{cite}
\usepackage{amsmath,amssymb,amsfonts}
\usepackage{graphicx}
\usepackage{algorithm,algorithmic}
\usepackage{hyperref}
\hypersetup{hidelinks=true}
\usepackage{textcomp}
\usepackage{booktabs}
\usepackage{multirow}
\def\BibTeX{{\rm B\kern-.05em{\sc 
\usepackage{makecell}
i\kern-.025em b}\kern-.08em
    T\kern-.1667em\lower.7ex\hbox{E}\kern-.125emX}}
\newcommand{\proc}[1]{\ifmmode\mbox{\textsc{#1}}\else\textsc{#1}\fi}

\newcommand{\htwin}{\proc{hePaTwin}}

\begin{document}
\title{A Physiology-Informed Digital Twin Framework for Simulating Liver Health Progression}
\author{Sumaiya Afroz Mila, and Sandip Ray, \IEEEmembership{Senior Member, IEEE}
\thanks{ Sumaiya Afroz Mila is with the Department of Electrical and Computer Engineering, University of Florida, Gainesville, FL 32611 USA (e-mail: mila.s@ufl.edu). }
\thanks{Sandip Ray is with the Department of Electrical and Computer Engineering, University of Florida, Gainesville, FL 32611 USA (e-mail: sandip@ece.ufl.edu). }
}

\maketitle

\begin{abstract}
We present a physiology-informed digital twin of the human liver designed for longitudinal simulation of liver function and early-stage disease progression. The model, referred to as $\htwin$, integrates key hepatic processes, including carbohydrate, lipid, and protein metabolism, bilirubin conjugation, bile production, and detoxification, within a unified systems-level framework to generate clinically observable biomarker trajectories. Unlike purely data-driven approaches, $\htwin$ incorporates mechanistic representations of liver physiology and patient-specific inputs such as diet, activity, and baseline biomarkers to simulate disease evolution over time. To ensure consistency with clinical progression patterns, we introduce a stage-transition-driven calibration mechanism that aligns simulated outputs with population-level biomarker distributions across disease stages, including NAFLD, fibrosis, and cirrhosis. Validation using the NIDDK NAFLD dataset demonstrates that $\htwin$ produces longitudinal biomarker estimates within clinically acceptable ranges and can forecast trajectories over multi-year horizons. Furthermore, simulated biomarkers retain sufficient clinical signal to support downstream NASH detection with competitive performance relative to models using ground-truth laboratory data. These results highlight the potential of  physiology-informed digital twins for personalized, non-invasive  diagnosis and prediction of organ health in general and liver health monitoring in particular.
\end{abstract}


\section{Introduction}
\label{sec:introduction}

\footnotetext{This work has been submitted to the IEEE for possible publication. Copyright may be transferred without notice, after which this version may no longer be accessible.}

\IEEEPARstart{L}{iver} diseases, particularly nonalcoholic fatty liver disease (NAFLD), nonalcoholic steatohepatitis (NASH), continue to be a major global health concern, with an increasing occurrence worldwide. These conditions typically progress silently from NAFLD to fibrosis, cirrhosis, and eventually hepatocellular carcinoma (HCC). The liver performs critical functions such as detoxification, metabolism, bile production, and protein synthesis. Despite being a vital organ, liver lacks nerve endings, making early-stage liver damage asymptomatic. By the time clinical symptoms manifest, often at the cirrhosis stage, the disease has already advanced, and treatment options focus on managing complications rather than reversing the condition.

The global burden of liver disease is profound, with cirrhosis alone causing approximately 1.48 million deaths globally in 2019, representing an 8.1\% increase from 2017 \cite{b1}. While the decline in hepatitis-related cirrhosis due to antiviral treatments has been notable, liver diseases caused by alcohol consumption and NAFLD are on the rise, particularly in populations with unhealthy dietary and lifestyle habits. Importantly, NAFLD is no longer confined to obese populations; more than 10\% of people with NAFLD in the U.S. are non-obese \cite{b2}. Despite its high prevalence, early-stage liver diseases like NAFLD, NASH and fibrosis are often underdiagnosed, underlining the critical need for early, non-invasive, accessible and continuous liver health monitoring solutions for better disease management or early detection\cite{milamasc}.

Artificial intelligence (AI) and machine learning (ML) technologies offer promising tools for early disease detection, including liver-related disorders. However, the effectiveness of these technologies depends heavily on the availability of robust training datasets. Most publicly available liver disease datasets are focused on advanced stages, such as cirrhosis, while intermediate stages like NAFLD, NASH and fibrosis are underrepresented or absent. Additionally, very few studies have systematically compared AI-driven approaches with physiology-informed models for continuous liver function simulation to detect early-stage diseases.


To address this gap, we present $\htwin$ (Liver P\underline{\textbf{h}}ysiology Mod\underline{\textbf{e}}ling from \underline{\textbf{Pa}}tient-Level Lifestyle Pattern in a Digital \underline{\textbf{Twin}} Framework for Precision Medicine), a physiology-informed digital twin of the human liver. $\htwin$ translates core hepatic physiological processes into computational modules, including carbohydrate, protein, fat, and cholesterol metabolism; protein synthesis; very low density lipoprotein (VLDL) export; urea excretion; bile production and excretion; and bilirubin conjugation. Using patient-specific inputs such as dietary intake, baseline biomarker values, demographic information, and physical activity, the model simulates longitudinal liver function and estimates clinically measurable biomarkers (e.g., ALT, ALP, albumin, bilirubin) under varying lifestyle conditions across multi-year prediction horizons.

Unlike purely statistical forecasting models, $\htwin$ incorporates physiological modeling of liver functions so that the simulated biomarker trends remain grounded in biological processes. Instead of predicting isolated biomarker values from data-driven learning, the $\htwin$ model simulates how liver-related biomarkers evolve over time under patient-specific lifestyle patterns. This allows the framework to produce longitudinal biomarker trajectories that reflect changes in liver health over time rather than producing static estimates. The proposed $\htwin$ framework contributes to the emerging paradigm of AI-enabled digital twins for personalized healthcare by integrating physiology-informed simulation with data-driven validation for early disease monitoring. By modeling core hepatic physiological processes and linking them with patient-specific lifestyle inputs, the digital twin enables individualized simulation of biomarker trajectories and disease progression under varying behavioral conditions. This capability supports anticipatory health monitoring and provides a computational foundation for decision-support systems that can assist clinicians and patients in understanding potential long-term liver health outcomes. Through this combination of personalized physiological modeling and clinically grounded validation, $\htwin$ advances the use of digital twin technologies for precision medicine and early-stage disease detection.

$\htwin$ is validated using the NIDDK NAFLD dataset through two strategies. First, $\htwin$ simulated biomarker trajectories are compared against longitudinal ground-truth clinical measurements to evaluate estimation accuracy across multiple follow-up visits. Second, to demonstrate the precision-medicine potential of the digital twin, we assess whether $\htwin$-generated biomarker trajectories preserve sufficient clinical signal to support downstream NASH detection. The detection performance obtained using estimated biomarkers is compared with baseline models trained on ground-truth data and with existing results reported on the same dataset.

The paper makes the following important contributions.

\begin{itemize}
    \item We develop a novel physiology-informed digital twin $\htwin$ that simulates key liver functions and estimates biomarker trajectories based on individual patient inputs, including diet, baseline biomarkers, physical profile, and activity patterns.

    \item We provide systematic validation of longitudinal biomarker estimation by comparing $\htwin$-generated outputs with ground-truth clinical data, demonstrating the model’s ability to reflect liver function progression in early-stage diseases such as NASH.

    \item We demonstrate $\htwin$'s capability to simulate biomarker evolution and disease trajectories under specified lifestyle conditions, enabling personalized, scenario-based evaluation of liver health trends.


    \item We demonstrate the feasibility of using $\htwin$-estimated biomarker trajectories for early-stage liver disease detection and continuous non-invasive monitoring of liver health.
\end{itemize}


The rest of the paper is structured as follows. We discuss related research in detecting liver disease progression and Non-Alcoholic Fatty Liver Disease in Section \ref{sec:related}.   In Section \ref{sec:background}, we present a detailed background on selected hepatic functions modeled in the proposed digital twin framework.  We then introduce a high-level overview of the design philosophy of $\htwin$, including the physiology-to-digital abstraction of hepatic functions and the overall validation strategy (Section \ref{sec:method}) and describe the detailed modeling, implementation, and computational modules underlying the framework (Section \ref{sec:designdetail}). We present and interpret the validation results in Section \ref{sec:validatehepatwin}, demonstrating the feasibility of the proposed $\htwin$ model.  We conclude in Section \ref{sec:concl}.

\section{Related Works}
\label{sec:related}

\subsection{Liver Disease Progression and Detection}

Liver disease progresses through various stages, beginning with hepatic fat accumulation, also known as Non-Alcoholic Fatty Liver Disease (NAFLD), which is followed by inflammation, fibrosis, tissue scarring, and ultimately cirrhosis and hepatocellular carcinoma (HCC). As the liver lacks sensory nerves, early-stage liver disease is often asymptomatic, resulting in delayed diagnosis and treatment. Early detection of NAFLD is crucial since it is a reversible condition, and appropriate lifestyle changes and interventions can halt its progression. However, once the liver damage reaches later stages such as fibrosis or cirrhosis, treatment options become limited and primarily focus on disease management rather than cure [5].

Traditional diagnostic methods for liver diseases, such as liver biopsy, have limitations due to their invasive nature, high cost, and potential risks to the patient. Non-invasive imaging techniques, such as ultrasound, computed tomography (CT), magnetic resonance imaging (MRI), and X-ray, have been developed to assess liver health more safely and efficiently. Magnetic resonance elastography (MRE) and ultrasound elastography are commonly used for assessing liver fibrosis, whereas MRI-proton density fat fraction (MRI-PDFF) is used for detecting hepatic steatosis [5], [6]. Despite the advancements in imaging technologies, these methods are often reserved for high-risk or symptomatic patients and are inconsistently applied, limiting their potential for early detection and prevention of liver diseases [9], [10].

\subsection{Non-Alcoholic Fatty Liver Disease (NAFLD) Detection}

Research in NAFLD detection has largely focused on imaging technologies like MRI-PDFF and transient elastography (TE), which measure liver fat content with greater accuracy compared to traditional methods such as ultrasound or controlled attenuation parameter (CAP) measurements [11], [12]. MRI-PDFF has shown strong correlations with histologic liver fat measurements, making it an important tool for detecting hepatic steatosis in clinical settings. However, early detection of NAFLD remains challenging due to the reliance on indirect methods that infer liver fat content rather than directly diagnosing the condition [5].

Liver biopsy, while the gold standard for confirming NAFLD, is invasive and costly. As such, it is not routinely used for early detection in the general population, which limits the ability to identify patients at risk before the disease progresses to more severe stages. The development of less invasive, reliable biomarkers for early-stage NAFLD is still an area of active research. Additionally, imaging biomarkers like MRI-PDFF are more commonly used to evaluate patients already diagnosed with fibrosis or cirrhosis, rather than as a primary method for detecting early NAFLD.

\subsection{Digital Twin Models in Liver Disease}

Digital Twin (DT) models, which simulate real-time organ physiology, have emerged as a promising approach for improving the management of liver diseases. These models can integrate patient data, including biomarkers, age, gender, and lifestyle factors, to create personalized virtual liver profiles that simulate disease progression or responses to treatment. In hepatology, DT models are being explored for use in predictive care and early intervention by tracking biomarker trends over time. By providing patient-specific simulations, DT models can offer valuable insights into liver function, aiding in the detection of early deviations from healthy trends before critical thresholds are crossed [13].

While most existing liver-focused DT models have been used for pharmacokinetic modeling or surgical planning, they remain limited in their ability to monitor disease progression in real-time. For instance, virtual liver platforms have been used to assess drug-induced liver injury [14], predict post-hepatectomy outcomes [15], and model liver regeneration at the cellular level [16]. 
Despite these advances, physiology-informed digital twin approaches for stage-specific biomarker estimation under data scarcity conditions remain limited. TwinScan, a partial digital twin framework, was developed to address this challenge by modeling stage-specific liver biomarker ranges using a hybrid strategy. The framework combined a machine learning–based mapping approach trained on cirrhosis-stage clinical data and healthy references with a physiology-informed digital abstraction of the bilirubin conjugation pathway. The model incorporated patient-level factors such as age and gender to enhance biological plausibility and personalization. While TwinScan demonstrated the potential of integrating mechanistic modeling with data-driven methods to generate clinically consistent biomarker estimates—particularly for intermediate disease stages. However, it remained limited to a partial pathway representation and did not model integrated liver functions or longitudinal state transitions across disease stages\cite{milatwinscan}.

\begin{figure*}[ht]
\centering
\begin{tabular}{cc}
\begin{minipage}{0.45\textwidth}
    \centering
    \includegraphics[width=\linewidth]{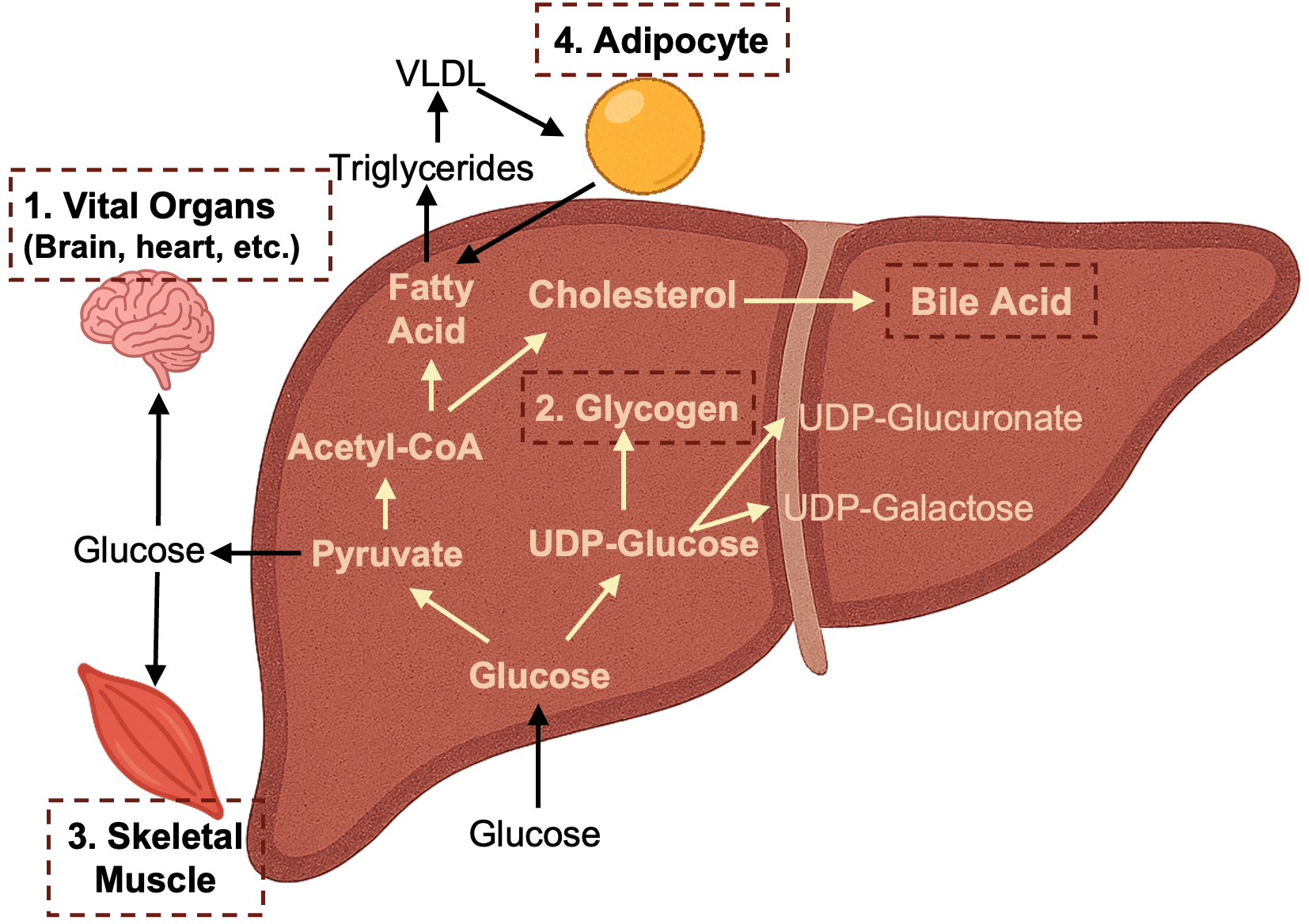}
\end{minipage} &
\begin{minipage}{0.45\columnwidth}
    \centering
    \includegraphics[width=\linewidth]{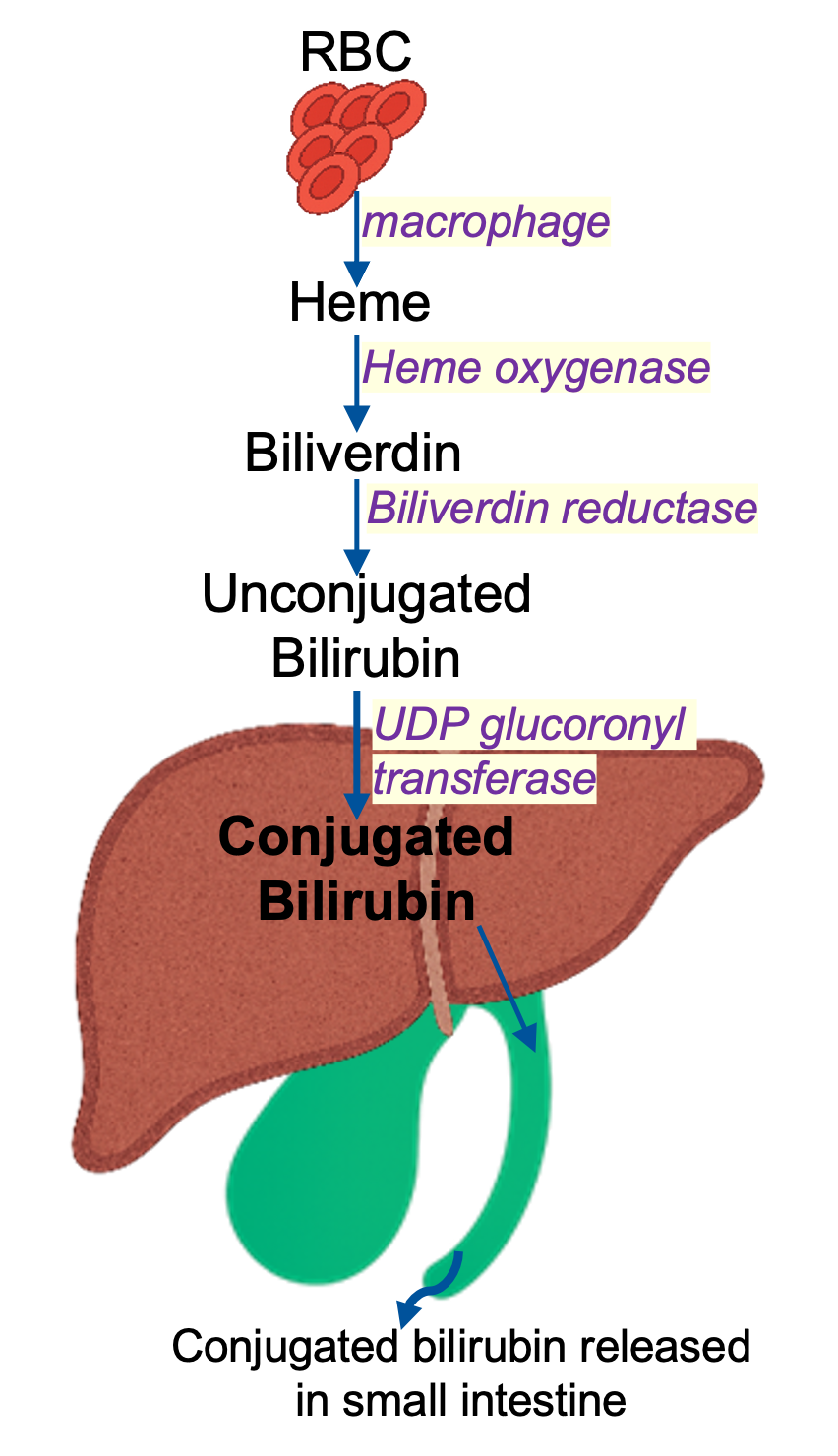}
\end{minipage}  \\
(a) Physiology of Glucose Metabolism & (b) Bilirubin Conjugation Pathway  \\
\end{tabular}

\caption{Simplified representation of selected liver physiological processes.}
\label{fig:bili_glucose_physiology}
\end{figure*}

\section{Background: Relevant Liver Physiology}
\label{sec:background}
\subsection{Overview of Liver Physiology (simplified) and Inter-Organ Communication}\label{subsec:overallphysiology}

The liver is a complex organ that performs a range of essential physiological functions, including metabolic homeostasis through carbohydrate, protein, and lipid metabolism; bile production; bilirubin conjugation; detoxification; and plasma protein synthesis (primarily albumin). Nutrient-rich blood from the gastrointestinal tract is delivered to the liver via the portal vein, where macronutrients are processed, stored, or redistributed based on systemic demand. The liver continuously communicates with peripheral organs such as the gallbladder, skeletal muscle, adipose tissue, and pancreas to regulate bile production and excretion, glucose availability, lipid storage, and insulin sensitivity. Through these interactions, the liver acts as a central metabolic hub that coordinates energy utilization and storage across the body.

In human physiology, this nutrient-rich blood enters the liver through the portal circulation, where incoming macronutrients are processed and either used for immediate energy, stored for future use, or redistributed to other organs according to systemic demand. In parallel, the liver performs important detoxification functions, including bile excretion and urea synthesis, contributing to waste removal through bile flow, urine, and stool \cite{kandalgaonkar2024digestive, solomando2022microplastic}.

These hepatic processes are regulated through continuous communication with peripheral organs. Signals from the pancreas convey information about blood glucose levels and insulin activity, which directly influence glucose metabolism, glycogen storage, fat synthesis, and ATP production \cite{lopez2022contribution}. Skeletal muscle communicates its glycogen storage status, reflecting conditions of glucose surplus or deficit \cite{guo2023metabolic}. Adipose tissue provides information regarding fat storage resulting from excess glucose converted in the liver and exported as very-low-density lipoproteins (VLDL) \cite{gilani2024adipose}. Communication with the gallbladder represents bile production, storage, and flow rate, which are essential for lipid digestion and bilirubin excretion \cite{IQWiG2021}.

Together, these interactions form a simplified but physiologically meaningful representation of liver-centered metabolic regulation with continuous feedback from peripheral organs. Fig.~\ref{fig:liverarch}(a) provides a high-level overview of simplified liver physiology and its interactions with other organs.

\subsection{Carbohydrate (Glucose) Metabolism Pathway}\label{subsec:carbohydrate_metabolism_physiology}

Carbohydrate metabolism is a primary
pathway for energy regulation in the human body and plays a central role in liver physiology. Glucose and fructose derived from dietary carbohydrate intake serve as the primary sources of energy. These macronutrients reach the liver through portal circulation, where they are processed and distributed according to immediate and anticipated energy demands. A fraction of incoming glucose is immediately allocated to meet the energy requirements of vital organs such as the brain, heart, and kidneys, while the remaining portion is directed toward storage or further metabolic processing. During carbohydrate metabolism, a portion of energy is inherently lost due to biochemical conversion inefficiencies.

Once immediate energy requirements are met, remaining glucose is stored as glycogen in the liver for later use during fasting or overnight periods. When hepatic glycogen storage approaches capacity, additional glucose is taken up by peripheral tissues, including skeletal muscle, where it is stored as glycogen for local energy use. Both the liver and skeletal muscle have finite glycogen storage limits \cite{Jensen2011Glycogen,Wasserman2009FourGrams}. When these limits are exceeded under conditions of sustained glucose surplus, excess glucose is directed toward de novo lipogenesis (DNL), contributing to fat accumulation in the liver or transport to adipose tissue as very-low-density lipoprotein (VLDL) \cite{glucosemetabolism}. This physiological sequence primarily occurs under conditions of dietary glucose surplus.
Conversely, during glucose deficit states, the body mobilizes alternative energy sources. Energy requirements are first met through available carbohydrates and lipids, and when these sources are insufficient, protein metabolism contributes to energy production \cite{veldhorst2009gluconeogenesis}. As dietary energy becomes inadequate, the body sequentially mobilizes stored energy reserves, beginning with hepatic glycogen, followed by increased reliance on adipose tissue stores \cite{Sanvictores2025Fasting}. Fig.~\ref{fig:bili_glucose_physiology}(a) illustrates the physiological pathway of glucose metabolism in the liver.

\subsection{Bilirubin Conjugation Pathway}\label{subsec:bili_physiology}

Bilirubin metabolism is another important physiological function of the liver associated with red blood cell turnover and waste removal. In human physiology, unconjugated bilirubin, which is water-insoluble, is continuously produced as a waste byproduct of old red blood cell (RBC) turnover and transported to the liver bound to albumin. Within hepatocytes, unconjugated bilirubin is converted into conjugated (direct) bilirubin, which is water-soluble and subsequently excreted into bile and transported through the biliary system for elimination from the body \cite{BROWN2017412}. Impairments in hepatocyte function, inflammation, or bile flow can disrupt this conjugation and excretion process, leading to bilirubin accumulation in circulation, which is clinically reflected by elevated total and direct bilirubin levels \cite{ramirez2024multifaceted}, \cite{Kalakonda2022Bilirubin}. Fig.~\ref{fig:bili_glucose_physiology}(b) illustrates the physiological bilirubin conjugation pathway.

\section{Methodology Overview --- $\htwin$ Design and Validation }\label{sec:method}


$\htwin$ is a physiology-driven liver digital twin designed to simulate key liver functions using patient-specific information, such as dietary intake, demographic and physical profile, activity level, and baseline liver biomarker values. The model captures core hepatic processes—such as metabolism, protein synthesis, bilirubin conjugation, bile production, and detoxification, while incorporating contextual interactions with other organs, including the gallbladder, pancreas, skeletal muscle, and adipose tissue. Based on the input information, $\htwin$ performs the core functions and generates updated liver biomarker values (ALT, ALP, albumin, and bilirubin) that reflect the evolving physiological state of the liver over time.

Each simulation instance represents longitudinal changes in liver health over a configurable number of days, where the duration can be dynamically defined based on the specified dietary and lifestyle conditions. The resulting simulated biomarkers are subsequently used for further analysis, including disease stage assessment such as NASH detection. By combining physiology-based simulation with patient context, $\htwin$ enables personalized modeling of liver disease progression and supports early, non-invasive monitoring in precision medicine settings.

\begin{figure*}
\centerline{\includegraphics[width=\textwidth]{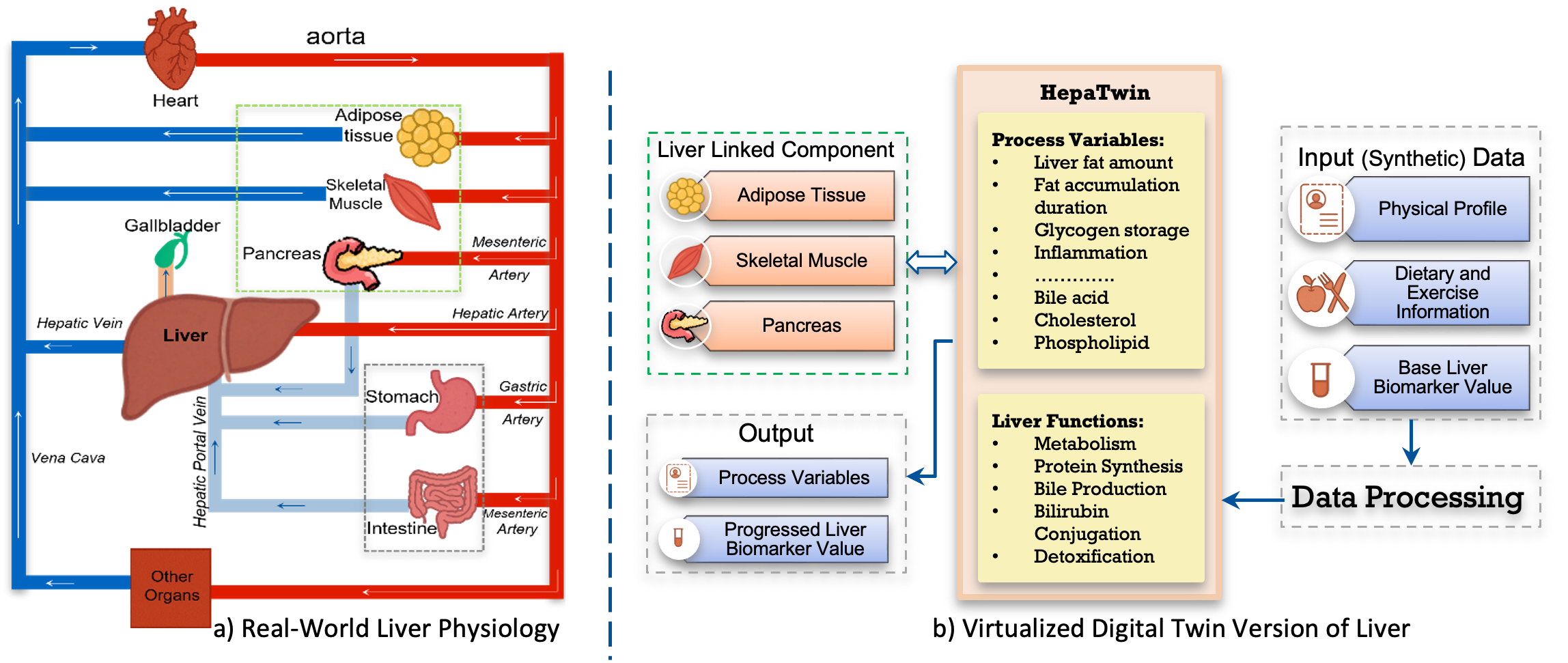}}
\caption{High-Level systemic overview of liver physiology and its virtualization: inter-organ
exchange of nutrients and metabolized products.}
\label{fig:liverarch}
\end{figure*}
\begin{table*}[t]
\caption{Hierarchical mapping of $\htwin$ components to liver functions}
\label{tab:hepatwin_modules}
\centering
\begin{tabular}{p{3.5cm} p{3.5cm} p{4.5cm} p{3cm}}
\toprule
\textbf{Main Component} & \textbf{Module} & \textbf{Submodule} & \textbf{High-level Liver Function} \\
\midrule
\multirow{3}{3.5cm}{Metabolism and Energy Homeostasis}

& Carbohydrate Metabolism 
& Energy expenditure; glucose regulation 
& Metabolism \\

& Lipid Metabolism 
& Fat storage; de novo lipogenesis; cholesterol regulation 
& Metabolism \\

& Protein Metabolism 
& Protein synthesis; amino acid regulation 
& Metabolism + Biosynthesis \\
\midrule
Detoxification and Waste Processing
& Bilirubin Conjugation 
& Bilirubin processing and clearance 
& Detoxification \\
\midrule
Biosynthesis, Production and Detoxification
& Bile Production and Excretion 
& Bile synthesis and transport 
& Biosynthesis + Detoxification \\
\bottomrule
\end{tabular}
\end{table*}

\subsection{Conceptual Design of $\htwin$}


$\htwin$ is designed to capture essential hepatic physiological roles at a systems level rather than reproducing detailed cellular biochemical pathways. In this work, we focus on hepatic functions that directly influence clinically observable liver biomarkers used in liver disease detection. These functions include carbohydrate metabolism and glycogen storage, lipid handling and fat accumulation, protein synthesis, bile production, bilirubin conjugation, and detoxification.

Fig.~\ref{fig:liverarch} presents a side-by-side conceptual illustration linking the simplified liver physiology described in Section~\ref{subsec:overallphysiology} with its abstraction into the digital space of $\htwin$. In the physiological view (Fig.~\ref{fig:liverarch}a), the simplified liver physiology described in Section~\ref{subsec:overallphysiology} is illustrated, highlighting the flow of nutrient-rich blood into the liver and its interactions with other organs involved in metabolic regulation. In the digital twin representation (Fig.~\ref{fig:liverarch}b), this physiological input is translated into structured inputs to $\htwin$, including dietary and exercise information, physical profile, and baseline biomarker values. After processing the input data to make them compatible with $\htwin$ input format, the processed data drives the core components of the digital twin, which maintains a set of process variables that track the evolving physiological state of the liver over time.

Bidirectional links between $\htwin$ and peripheral organ components, represented as "Liver Linked component" (green dashed block in Fig.~\ref{fig:liverarch}b), represent contextual physiological interactions rather than full mechanistic organ models. This design allows $\htwin$ to remain computationally tractable while preserving the influence of extra-hepatic factors that modulate liver metabolism and disease risk. We design $\htwin$ to model liver physiology by organizing hepatic functions into three interacting components:
\begin{itemize}
    \item Metabolism and Energy Homeostasis
    \item Detoxification and Waste Processing
    \item Biosynthesis and Production
\end{itemize}

Each of these components is implemented through computational submodules (as represented in Table~\ref{tab:hepatwin_modules}) that collectively drive longitudinal biomarker dynamics.

\subsection{$\htwin$ Architecture and Digital Abstraction}

The $\htwin$ framework consists of interconnected conceptual blocks that translate patient context into simulated liver outcomes. Patient-specific inputs are first processed to derive relevant physiological parameters. These parameters inform the $\htwin$ core model, which simulates liver functions through the modules and submodules, and updates internal process variables reflecting liver fat accumulation, glycogen storage, bile-related processes, protein synthesis status etc. Based on these evolving states, $\htwin$ generates simulated liver biomarker values corresponding to clinical measurements associated with each patient's health state.

This abstraction enables $\htwin$ to function as a virtual liver that evolves dynamically in response to diet, lifestyle and patient-profile, providing a longitudinal estimate of liver health rather than a static snapshot. Detailed mathematical formulations and update mechanisms of these processes are described in Section~\ref{sec:designdetail}.

\begin{figure}
\centerline{\includegraphics[width=\columnwidth]{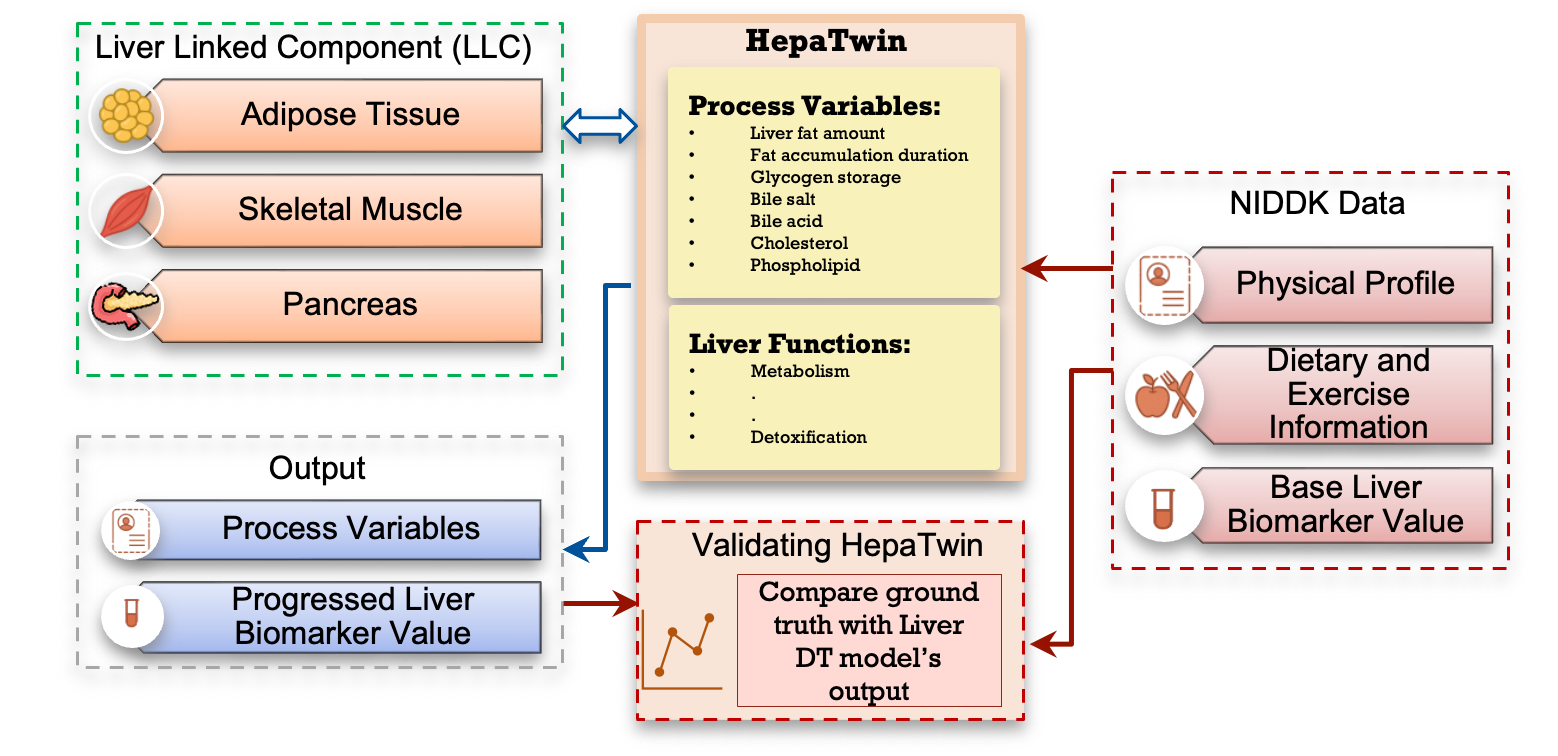}}
\caption{Validation Strategy I: Biomarker Estimation Accuracy through Ground Truth Biomarker Comparison.}
\label{fig:val1}
\end{figure}

\begin{figure*}
\centerline{\includegraphics[width=.9\textwidth]{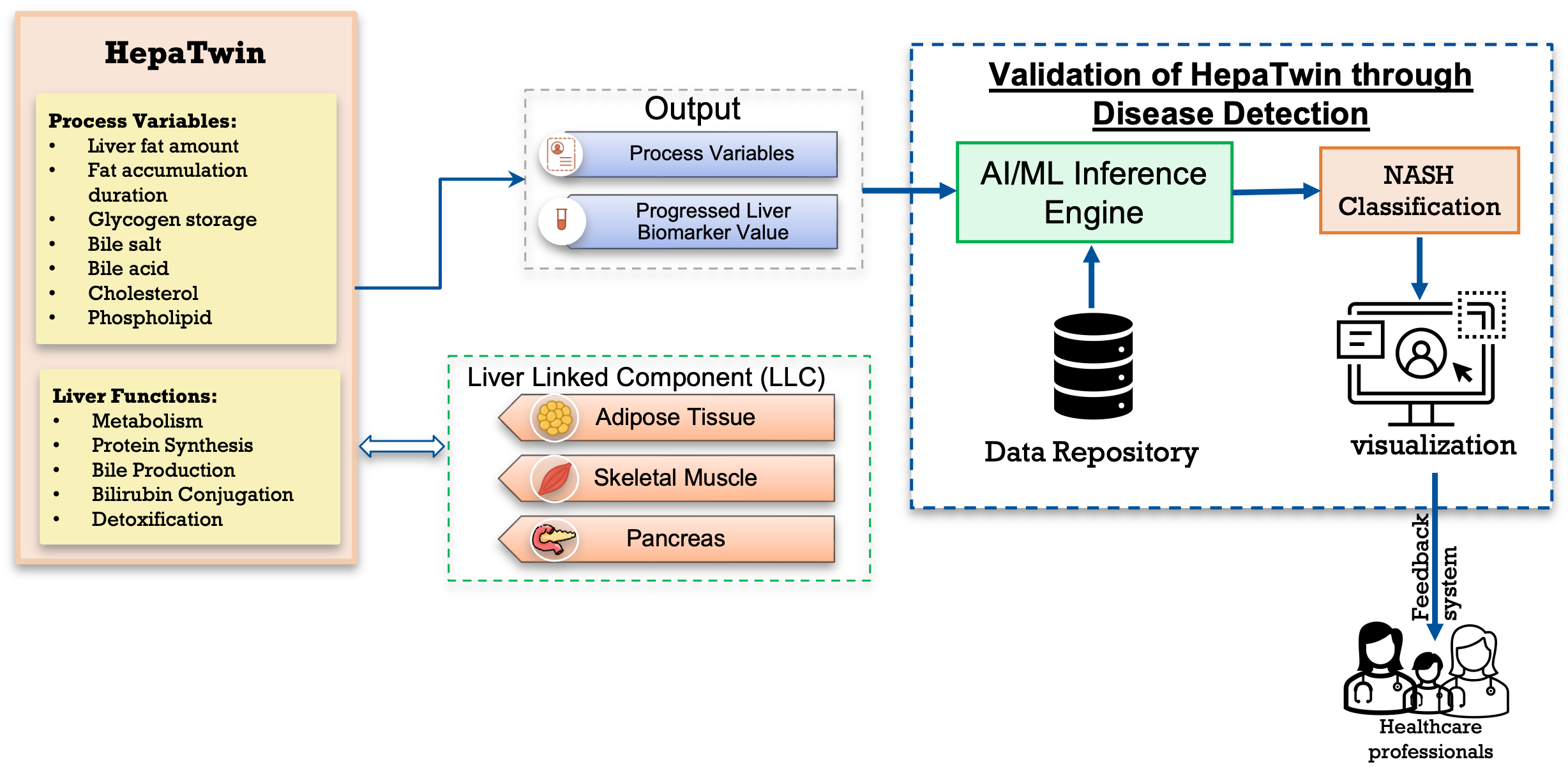}}
\caption{Validation Pathway II: NASH Classification via $\htwin$ Outputs.}
\label{fig:val2}
\end{figure*}

\subsection{Validation Strategy Overview}

To evaluate the reliability and practical utility of the $\htwin$ model, we employ two complementary validation strategies, illustrated in Fig.s~\ref{fig:val1} and ~\ref{fig:val2}.

\textit{Validation Strategy I} focuses on biomarker estimation accuracy by comparing $\htwin$-generated liver biomarker values with corresponding clinical ground truth measurements. This strategy assesses whether the physiology-driven simulation can closely reproduce observed biomarker trends across patient visits.

\textit{Validation Strategy II} evaluates downstream disease detection feasibility by using $\htwin$-simulated outputs as inputs to an independent AI/ML inference pipeline for NASH classification. This strategy evaluates whether $\htwin$'s simulated biomarkers preserve clinically meaningful patterns that support disease stage discrimination.

Together, these strategies evaluate both the physiological fidelity and clinical relevance of $\htwin$. Detailed dataset descriptions, experimental protocols, and quantitative results are presented in Section~\ref{sec:validatehepatwin}.

\subsection{Summary of Methodology}
In summary, we introduce a digital twin model of the liver, $\htwin$, that simulates liver function under patient-specific dietary and lifestyle conditions. By abstracting key hepatic processes into a computational framework and validating its outputs through direct biomarker comparison and downstream disease detection tasks, $\htwin$ provides a personalized, non-invasive approach for monitoring liver health and disease progression. This overview establishes the conceptual foundation for the detailed modeling and validation presented in the subsequent sections. 

\section{$\htwin$ DESIGN: FROM REAL-WORLD PHYSIOLOGY TO DIGITAL TWIN}
\label{sec:designdetail}

This section describes the detailed design of $\htwin$, including the translation of real-world liver physiology into a digital abstraction and the resulting $\htwin$ architecture. Building on the physiological background described in Section~\ref{sec:background}, we present the methodology used to convert key hepatic processes into a computational digital twin framework. Carbohydrate metabolism and bilirubin conjugation are selected as representative hepatic functions to demonstrate the $\htwin$ design, as they capture two complementary aspects of liver physiology: metabolic energy regulation and detoxification. These pathways are described in detail to illustrate the translation of liver physiological mechanisms into the digital twin framework. Other modules, protein metabolism, lipid metabolism, bile production and excretion, and urea synthesis are briefly summarized, as they follow the same modeling framework. The section concludes by describing the method used to estimate the longitudinal changes in biomarker values from baseline measurements.

\begin{figure}
\centerline{\includegraphics[width=.8\columnwidth]{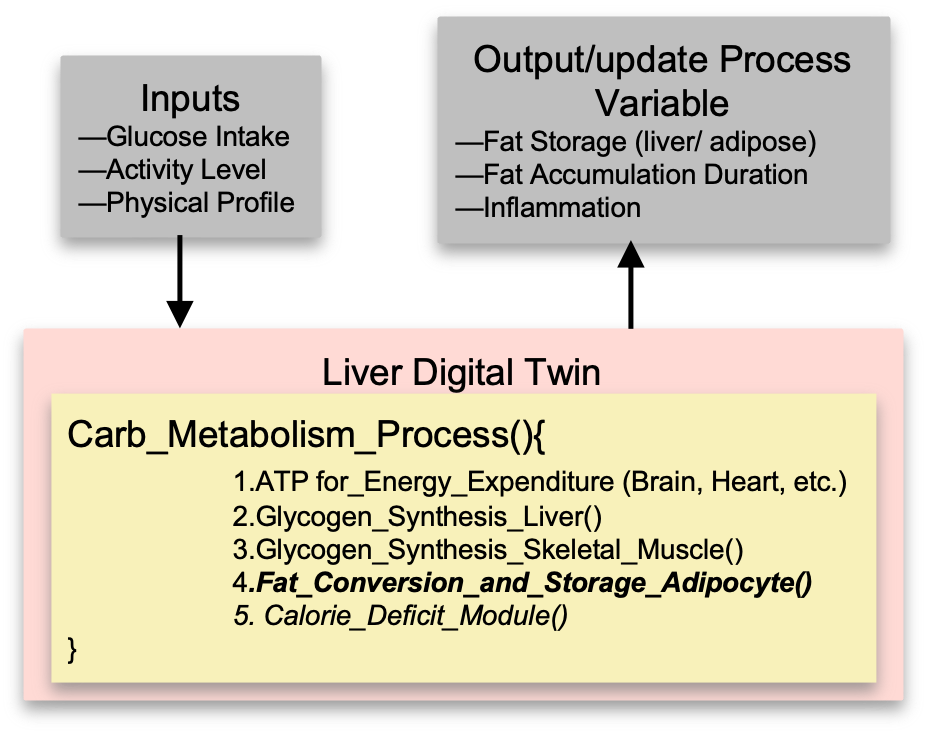}}
\caption{Digital Abstraction of Hepatic Glucose Metabolism in $\htwin$.}
\label{fig:glucosemetabolism}
\end{figure}



\subsection{Digital Abstraction of Liver Physiology in $\htwin$}

To translate the real-world physiology into a digital twin, the macronutrients arriving at the liver through the GI tract are represented using patient-specific dietary information, including total caloric intake and macronutrient composition (carbohydrates, proteins, and lipids). In human physiology, the liver continuously communicates with peripheral organs to regulate metabolic states such as fed and fasting conditions, caloric surplus or deficit, and overall energy demand. In $\htwin$, these physiological states are abstracted through a set of internal process variables initialized using each patient’s physical profile. For example, body mass index (BMI) and total daily energy expenditure (TDEE) are computed using age, sex, height, weight, and daily activity level, with TDEE estimated using the Mifflin–St Jeor equation\cite{macena2023estimates}. These variables provide a quantitative representation of systemic energy requirements and metabolic context, which guide subsequent liver function simulations. 

Baseline liver biomarker values are initialized using measurements collected during the screening visit. While a detailed discussion of data distribution and cohort characteristics from the NIDDK NAFLD Adult dataset is presented in Section~\ref{sec:validatehepatwin}, we briefly note here that each patient has multiple follow-up visits spaced approximately 48 weeks apart. These visits include a screening visit followed by longitudinal assessments at weeks 48, 96, 144, and 192. For this study, only patients with a minimum of two visits are selected, with screening biomarker values serving as baseline inputs to $\htwin$.

Using these patient-specific inputs, $\htwin$ simulates subsequent visit-level biomarker values by integrating dietary intake, physical profile, and inter-organ communication into its internal liver modules and process variables. Through this process, $\htwin$ models liver function by simulating hepatic metabolism, nutrient storage, redistribution, and detoxification mechanisms over time. This workflow forms the core mechanism through which liver physiology is translated into a digital simulation framework.

In the following subsections, we describe the digital implementation of two key hepatic functions---carbohydrate metabolism and bilirubin conjugation, illustrating how the physiological processes introduced in Section~\ref{sec:background} are translated into the $\htwin$ computational framework.


\subsection{Digital Abstraction of the Glucose Metabolism Pathway in $\htwin$}

The physiological behavior described in subsection~\ref{subsec:carbohydrate_metabolism_physiology} forms the basis of the glucose metabolism pathway we modeled in $\htwin$. In the digital twin, glucose metabolism is implemented as a dedicated carbohydrate (glucose) metabolism module, with glucose modeled as a subset of total carbohydrate intake. For modeling simplicity, this work focuses on the glucose metabolism pathway, while fructose metabolism is briefly discussed in a later subsection, recognizing that fructose follows distinct regulatory mechanisms but ultimately contributes to common hepatic energy utilization and storage processes. The glucose metabolism module consists of multiple submodules that abstract key physiological processes, including immediate energy expenditure, liver glycogen storage, skeletal muscle glycogen storage, fat conversion and storage in adipose tissue, and energy compensation under caloric deficit conditions.

Inputs to the glucose metabolism module include dietary glucose intake, activity level, demographic and physiological profile. These inputs are used to compute the caloric state and total daily energy expenditure (TDEE), estimated using the Mifflin–St Jeor equation\cite{macena2023estimates}. A tunable proportion of TDEE (approximately 65-85\%, based on typical dietary energy utilization patterns) is first allocated to immediate energy expenditure using dietary glucose\cite{volek2024expert}. Remaining glucose is then routed through glycogen storage submodules.

Liver glycogen storage is constrained to a maximum capacity of approximately 120 g, while skeletal muscle glycogen storage is limited to approximately 350–400 g\cite{Jensen2011Glycogen,Wasserman2009FourGrams}. These storage capacities are represented in the model as internal process variables that track hepatic and skeletal muscle glycogen storage and adipose fat accumulation over time. Once glycogen storage limits are reached, excess glucose is forwarded to the fat conversion submodule, where it contributes to fat accumulation in the liver or is transported to adipose tissue. Under caloric deficit conditions, control is transferred to a deficit compensation module, in which energy demands are sequentially met through hepatic glycogen depletion followed by adipose tissue stores.

Fig.~\ref{fig:glucosemetabolism} presents the digital implementation of the hepatic glucose metabolism pathway, highlighting how physiological processes are translated into computational modules within the $\htwin$ framework.

\begin{figure}
\centerline{\includegraphics[width=.5\columnwidth]{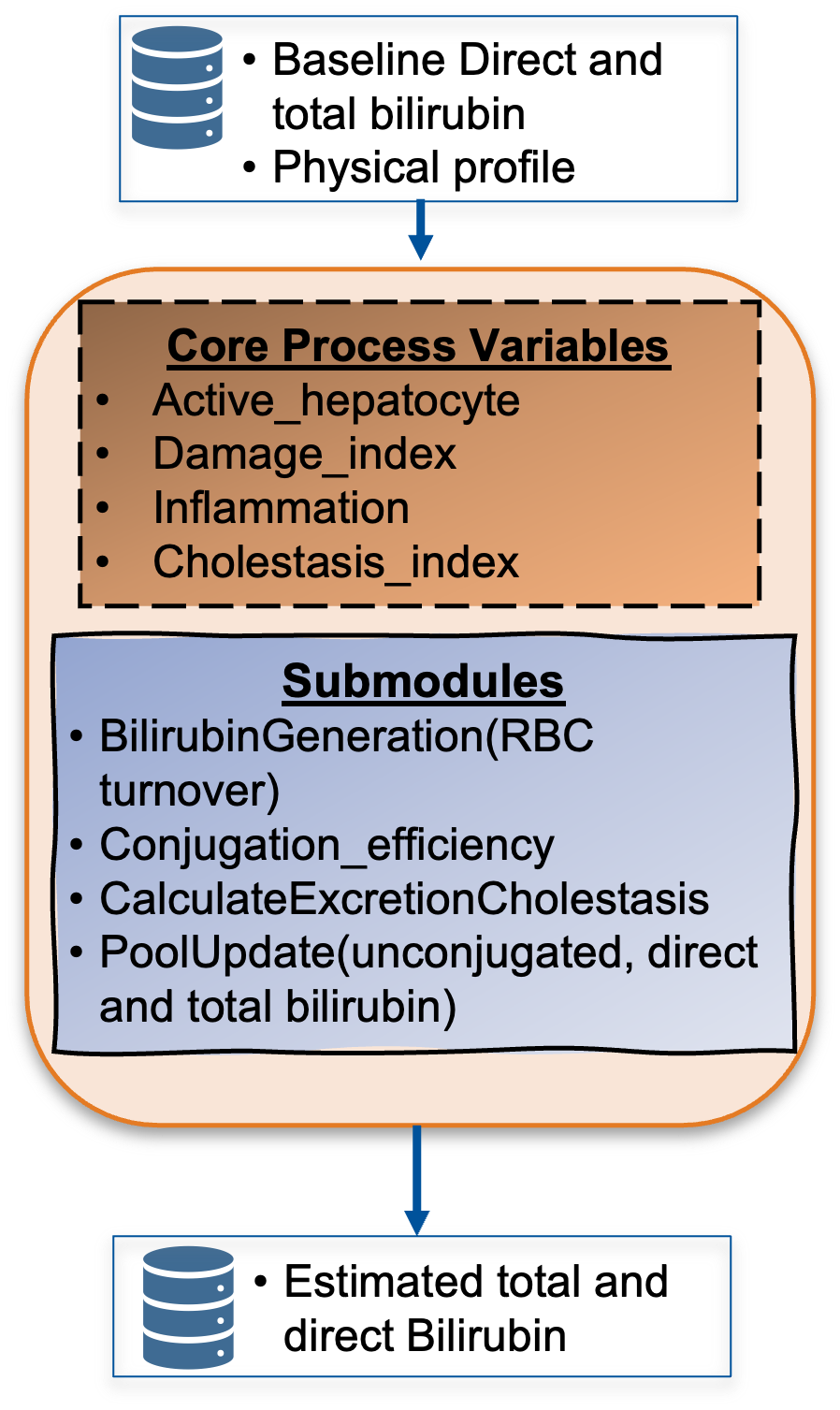}}
\caption{Digital Abstraction of the Bilirubin Conjugation Pathway in $\htwin$.}
\label{fig:bilirubin}
\end{figure}

\begin{algorithm}
\caption{Bilirubin Conjugation Pathway in $\htwin$}
\label{alg:bilirubin}
\begin{algorithmic}[1]

\STATE \textbf{Input Variables:} Baseline Bilirubin $B^{(0)}, B_d^{(0)}$, Gender $G$, Age $A$
\STATE \textbf{Process Variables:} Stage $D \in \{0,1,2,3,4,5\}$, hepatocyte\_capacity $c$, inflammation $i$, cholestasis $s$, damage $\delta$

\STATE \textit{// Physiological Modulators}
\STATE $\texttt{ox} \gets f_1(D,c,i)$ //oxidative stress
\STATE $\texttt{hem\_eff} \gets f_2(\texttt{ox},A,G)$ //Heme catabolism efficiency
\STATE $\texttt{hem\_rate} \gets f_3(D,c,i,A,G)$ //Hemolysis rate
\STATE $\texttt{UGT} \gets f_4(c,i,A,G)$ //UGT activity
\STATE $\texttt{conj\_cap} \gets f_5(\texttt{ox},A,G,\delta)$ //conjugation capacity

\STATE \textit{// Bilirubin Production}
\STATE $\texttt{B}_{\text{prod}} \gets f_6(\texttt{hem\_eff},\texttt{hem\_rate})$  //Unconjugated bilirubin production from RBC turnover

\STATE \textit{// Initialize Pools}
\STATE $B_{\text{tot}}^{(t)} \gets B^{(0)}$
\STATE $B_{\text{dir}}^{(t)} \gets B_d^{(0)}$
\STATE $B_{\text{unc}}^{(t)} \gets \max(0, B_{\text{tot}}^{(t)} - B_{\text{dir}}^{(t)})$

\STATE \textit{// Conjugation Efficiency}
\STATE $\eta \gets f_7(UGT,conj\_cap)$
\STATE $\eta \gets \max(0.05,\min(0.90,\eta))$

\STATE \textit{// Pool Update}
\STATE $B_{\text{in}} \gets B_{\text{prod}} + B_{\text{unc}}^{(t)}$
\STATE $B_{\text{conv}} \gets \eta \cdot B_{\text{in}}$
\STATE $B_{\text{unc}}^{(t+1)} \gets \max(0, B_{\text{in}} - B_{\text{conv}})$

\STATE \textit{// Cholestasis Retention}
\STATE $\rho \gets f_8(s)$
\STATE $\tilde{B}_d^{(t+1)} \gets 0.85 B_{\text{dir}}^{(t)} + \rho B_{\text{conv}}$

\STATE \textit{// Recompose Total}
\STATE $\tilde{B}^{(t+1)} \gets B_{\text{unc}}^{(t+1)} + B_{\text{dir}}^{(t+1)}$

\end{algorithmic}
\end{algorithm}

\subsection{Bilirubin Conjugation: Physiology-Informed Digital Abstraction}


In the digital twin, the baseline total and direct bilirubin levels are initialized with laboratory measurements obtained during the screening visit. Continuous production of unconjugated bilirubin from RBC turnover is modeled as a sustained influx process influenced by patient-specific physiological characteristics. Separate internal pools are maintained for total and direct bilirubin, allowing the unconjugated bilirubin component to be inferred dynamically. Conversion from unconjugated to conjugated bilirubin is governed by a set of physiology-informed process variables, including hepatocyte functional capacity, liver damage index, inflammation index, and cholestasis index. Conjugation efficiency is bounded within physiologically plausible limits to maintain numerical stability and prevent nonphysical saturation across disease stages. Fig.~\ref{fig:bilirubin} shows the digital abstraction of the Bilirubin Conjugation Pathway in $\htwin$, highlighting the core process variables and computational submodules.

Under normal conditions, most newly conjugated bilirubin is excreted via bile. However, when cholestasis is elevated in disease states, a larger proportion of conjugated bilirubin is retained in circulation, leading to increased direct bilirubin levels. This mechanism enables $\htwin$ to reproduce clinically observed patterns of conjugated, unconjugated, and total bilirubin with impaired bile excretion conditions.

Through this abstraction, $\htwin$ models bilirubin conjugation and clearance as a dynamic, state-dependent process influenced by hepatocyte capacity, inflammation, and cholestasis. This design enables the simulation of clinically observable bilirubin patterns while remaining computationally tractable and aligned with the system-level philosophy of the digital twin. Algorithm~\ref{alg:bilirubin} presents the mechanistic bilirubin conjugation workflow prior to stage-level calibration. The raw total bilirubin estimate 
$\tilde{B}^{(t+1)} $
 represents the mechanistic output of the bilirubin conjugation module prior to stage-level calibration. UGT activity is modeled as a function of hepatocyte capacity and inflammatory state; therefore, the conjugation efficiency 
$\eta$ is expressed in terms of UGT activity and effective conjugation capacity, without explicitly reintroducing the hepatocyte capacity ($c$) and inflammation variable ($i$), as their effects are already embedded within UGT.
\subsection{Brief Description of Other Modules}
In addition to the detailed pathways described earlier, $\htwin$ integrates several additional hepatic modules that follow the same physiology-informed abstraction framework. These modules are summarized below.
\paragraph*{Fructose Metabolism} Fructose metabolism in human physiology primarily occurs in the liver, where dietary fructose is transported via portal circulation and processed within hepatocytes. Unlike glucose, fructose metabolism bypasses key regulatory steps of glycolysis and is preferentially directed toward intermediates that contribute to energy production, glycogen synthesis, and de novo lipogenesis (DNL). As a result, excessive fructose intake is more strongly associated with hepatic fat accumulation\cite{schaefer2009dietary}.

In the digital twin, fructose is modeled as a distinct carbohydrate input when it is explicitly specified in the dietary intake data; otherwise, total carbohydrate intake is primarily treated as glucose for modeling simplicity. Fructose converges on shared downstream hepatic pathways governing energy utilization and lipid storage but is assigned a higher lipogenic weighting compared to glucose to reflect its stronger contribution to DNL. In the energy allocation hierarchy, glucose is prioritized before fructose for immediate energy expenditure\cite{lu2024effects}. When glucose availability is sufficient to meet energy demands, fructose-derived substrates are preferentially directed toward storage pathways, contributing to hepatic and adipose fat accumulation. Rather than modeling fructose-specific enzymatic reactions, $\htwin$ incorporates fructose within the carbohydrate metabolism module while preserving its physiologically relevant outcome, such as enhanced lipid synthesis keeping it consistent with the system-level abstraction of the digital twin.

\begin{figure*}
\centerline{\includegraphics[width=\textwidth]{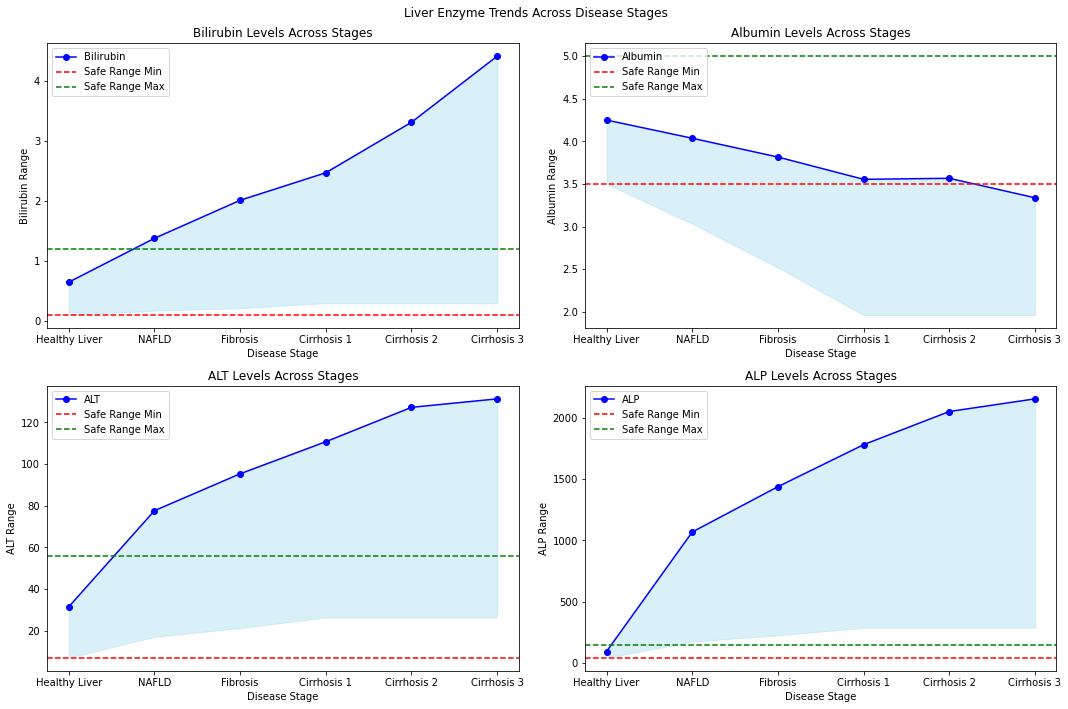}}
\caption{Distribution (estimated NAFLD, Fibrosis; clinical data Chirrohis stage 1, 2, 3; Literature reference healthy) of Biomarker Values across Disease Stages.}
\label{fig:progressfactor}
\end{figure*}

\paragraph*{Protein Metabolism} Protein metabolism in $\htwin$ is modeled to capture the liver’s role in amino acid regulation, protein synthesis, and energy compensation under nutrient-deficit conditions. Dietary protein intake is represented as amino acid availability to the liver, where a portion is allocated toward hepatic protein synthesis, including albumin production\cite{THALACKERMERCER20071734}. When energy demands cannot be met through carbohydrate and lipid metabolism, protein metabolism contributes to energy production through amino acid catabolism and gluconeogenesis, with the high thermal effect of protein accounted for during this conversion process \cite{paulusma2022amino}. Under sustained protein deficiency or metabolic stress, reduced protein synthesis is reflected in downstream biomarker changes, enabling $\htwin$ to model clinically observable effects associated with impaired hepatic protein metabolism.

\paragraph*{Lipid Metabolism} Lipid metabolism in $\htwin$ models the hepatic handling of dietary lipids, including their role in supporting energy production, fat storage, mobilization, and transport. Dietary lipid intake and excess glucose-derived substrates contribute to hepatic fat accumulation through DNL and lipid storage processes\cite{ZOU2023275},\cite{Nguyenliver}. The module tracks lipid storage dynamics in the liver and adipose tissue and incorporates lipid mobilization during caloric deficit states to support energy requirements. Through this abstraction, $\htwin$ captures the progression of lipid accumulation and redistribution that underlies metabolic dysfunction and fatty liver disease.

\paragraph*{Bile Production, Flow and  Excretion}Bile production in $\htwin$ models the liver’s role in synthesizing bile acids, incorporating conjugated bilirubin into bile, and regulating biliary flow dynamics under varying physiological states. Bile formation is driven primarily by lipid metabolism and hepatocyte functional capacity\cite{ahmed2022functional}. Dietary lipids (cholesterol) stimulate gallbladder contraction and bile secretion into the intestine, where bile acids represent the principal functional component of bile for lipid emulsification\cite{Hundt2022BileSecretion}. Conjugated bilirubin, generated through hepatic bilirubin metabolism, is secreted as a component of bile and contributes to the physiological excretion of heme degradation products\cite{Hundt2022BileSecretion}. Rather than modeling detailed enzymatic reactions of bile acid synthesis, $\htwin$ abstracts bile production as a flux-based process that reflects substrate availability and hepatocyte capacity.

The module maintains a process variable representing the bile pool's  temporary storage and availability prior to excretion. Bile flow is regulated by hepatocyte functional capacity and inflammatory state. Under healthy conditions, adequate capacity supports consistent bile drainage. However, reduced hepatocyte capacity and elevated inflammation increase a cholestasis index, representing impaired bile flow\cite{cholestasis}. This index modulates the fraction of bile excreted during each simulation interval, enabling gradual accumulation of conjugated bilirubin under impaired states and enhanced clearance when function is restored.

By linking bile flow to both hepatocyte capacity, cholestasis, and inflammation, $\htwin$ captures clinically relevant dynamics such as impaired biliary drainage and conjugated bilirubin retention. The model exposes bile production rate, bile flow rate, and cholestasis index as process markers that interact with the bilirubin module, ensuring physiologically consistent coupling between bile physiology and serum bilirubin trends while preserving the system-level abstraction of the digital twin framework.

\subsection{Biomarker Trajectory Estimation}
After execution of the physiological modules, $\htwin$ updates its internal process variables, including inflammation index, active hepatocyte fraction, damage index, cholestasis index, total hepatic fat amount, hepatic fat duration, hepatic fat delta, adipose fat storage, and the post-physiology liver state. These updated process variables reflect the cumulative effect of metabolic, detoxification, and storage mechanisms simulated during the visit interval. Following this internal state update, control is transferred to the biomarker trajectory estimation module. While the physiology modules produce mechanistic biomarker estimates based on simulated hepatic function, population-level disease stage transition patterns are incorporated through a stage-transition–driven calibration step.  To construct stage-consistent biomarker reference distributions, we utilize the Cirrhosis Patient Survival dataset, which provides cirrhosis stage–specific biomarker values (Stages 1--3)\cite{cirrhosis_patient_survival_prediction_878}. Healthy reference ranges are integrated with this dataset, sample sizes are balanced, and a statistical regression mapping is developed to estimate biomarker distributions for intermediate disease stages, including NAFLD and Fibrosis. Fig.~\ref{fig:progressfactor} illustrates the safe ranges for each biomarker (bilirubin, ALT, ALP, albumin), the observed distributions for healthy and cirrhosis stages, and the estimated distributions for intermediate stages\cite{milatwinscan}.  Once the physiology simulation is completed, the pre-physiology disease state $D_{\text{pre}}$
 and post-physiology disease state $D_{\text{post}}$ are used to compute a progress factor 
$p = f\_{\text{prog}}(D\_{\text{pre}}, D\_{\text{post}})$
 using the stage-specific biomarker mapping derived from the dataset. This progress factor represents the direction and magnitude of disease progression or recovery over the simulated interval.

The final biomarker estimate is obtained by applying a baseline-referenced drift correction to the mechanistic output:
\[
B^{(t+1)} = \tilde{B}^{(t+1)} + (p - 1)\, B^{(t)}
\]

Here
$\tilde{B}^{(t+1)}$ is the mechanistic estimate generated by the physiology module in Algorithm~\ref{alg:bilirubin},
${B}^{t}$ is the baseline biomarker value provided as input to $\htwin$, and
$p$  is the stage-transition–driven progress factor.
When $D_{\text{pre}}$=$D_{\text{post}}$, $p=1$, and the final estimated value equals the mechanistic output. When disease severity increases, $p>1$, producing an upward drift relative to baseline; conversely, when disease severity improves, $p<1$, resulting in a downward adjustment.

Algorithm~\ref{alg:bili_progress_factor} presents the detailed procedure for estimating updated total and direct bilirubin values. Other biomarkers (ALT, ALP, and albumin) are updated following a similar stage-transition–based calibration framework, ensuring consistency between simulated hepatic physiology and population-level disease stage transition patterns.

\begin{algorithm}
\caption{Estimating Biomarker (Bilirubin) Value After Physiology Simulation using Progress-Factor Calibration}
\label{alg:bili_progress_factor}
\begin{algorithmic}[1]

\STATE \textbf{Input Variables:} Pre-physiology stage $D_{\text{pre}}$, post-physiology stage $D_{\text{post}}$
\STATE \hspace{1.25em} Baseline biomarker states $B^{(0)}, B_d^{(0)}$
\STATE \hspace{1.25em} Trend table $\texttt{BILI\_TRENDS}$ indexed by stage (for bounds/statistics)

\STATE \textbf{Process Variable:} Post-physiology (mechanistic) outputs $\tilde{B}^{(t+1)}, \tilde{B}_d^{(t+1)}$

\STATE \textit{// Compute progress factor from stage transition}
\STATE $p \gets f_{\text{prog}}(D_{\text{pre}}, D_{\text{post}})$
\STATE $p \gets \max(p_{\min}, \min(p_{\max}, p))$ \hfill //  clamp for stability

\STATE \textit{// // Apply stage-transition baseline drift correction}
\STATE $B^{(t+1)} \gets \tilde{B}^{(t+1)} + (p-1){B}^{(0)}$ 
\STATE $B_d^{(t+1)} \gets \tilde{B}_d^{(t+1)} + (p-1){B}_d^{(0)}$ 

\STATE \textit{// Enforce non-negativity}
\STATE $B^{(t+1)} \gets \max(0, B^{(t+1)})$
\STATE $B_d^{(t+1)} \gets \max(0, B_d^{(t+1)})$

\STATE \textit{//  stage-consistent clamping using trend bounds}
\STATE $(B_{\min}, B_{\max}, B_{\mu}) \gets \texttt{BILI\_TRENDS}[D_{\text{post}}]$
\STATE $B^{(t+1)} \gets \min(B_{\max}, \max(B_{\min}, B^{(t+1)}))$
\STATE $B_d^{(t+1)} \gets \min(0.85B_{\max}, \max(0.05B_{\min}, B_d^{(t+1)}))$

\STATE \textit{// Maintain pool consistency (from tracked unconjugated)}
\STATE $B_u^{(t+1)} \gets \max(0, B^{(t+1)} - B_d^{(t+1)})$

\STATE \textbf{return} $B^{(t+1)}, B_d^{(t+1)}, B_u^{(t+1)}$

\end{algorithmic}
\end{algorithm}

\begin{figure*}
\centerline{\includegraphics[width=.96\textwidth]{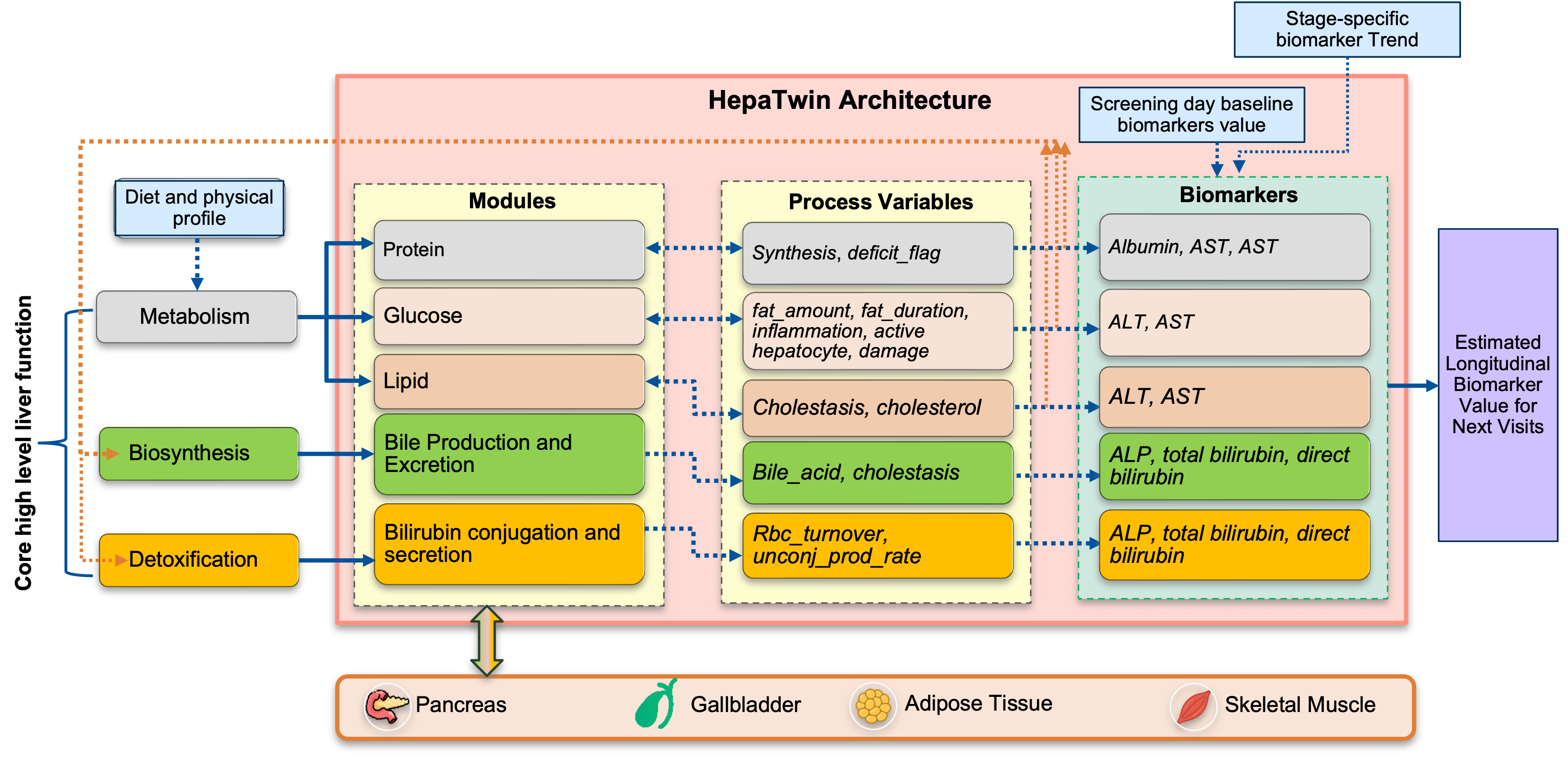}}
\caption{Directed architecture inside $\htwin$ illustrating physiology-to-digital abstraction, showing the left-to-right propagation from hepatic functions to computational modules, intermediate process variables, and observable biomarker outputs.}
\label{fig:hepatwininternalarchitecture}
\end{figure*}

\subsection{Summary of Digital Abstraction of Liver Physiology}
Fig.~\ref{fig:hepatwininternalarchitecture} illustrates the directed network structure of $\htwin$, summarizing how hepatic physiological functions are abstracted into computational modules and translated into measurable biomarker outputs. Core liver functions—metabolism, detoxification, and biosynthesis are decomposed into dedicated modules (e.g., glucose, protein, and lipid metabolism; bile synthesis and flow regulation; conjugation capacity and RBC turnover), and corresponding submodules that regulate intermediate process variables, such as fat accumulation, fat storage duration, inflammation index, cholestasis index, and conjugation efficiency, which propagate forward through the digital twin architecture. These process variables are integrated with baseline biomarker states and stage-constrained trend functions to collectively regulate downstream biomarker dynamics, including ALT, AST, ALP, albumin, total bilirubin, and direct bilirubin, producing longitudinal biomarker estimates for subsequent visits.
The left-to-right structure emphasizes causal signal flow from hepatic physiological processes to digital state variables and ultimately to clinically observable biomarkers, preserving system-level abstraction while maintaining physiological interpretability. Unlike purely data-driven models that learn statistical associations between inputs and outputs, $\htwin$ explicitly encodes physiological mechanisms and causal dependencies within its computational architecture.

Solid arrows represent hierarchical decomposition of core functions into computational modules and submodules. Dashed arrows indicate functional dependencies between components. A single-headed dashed arrow (e.g., from bile production to bile acid synthesis) denotes directional dependency, where the process variable is regulated by the module but does not directly influence the module in return. In contrast, bidirectional dashed arrows (e.g., between glucose metabolism and fat accumulation/inflammation) represent feedback coupling. For example, high glucose intake increases hepatic fat accumulation; sustained fat accumulation elevates inflammation; and elevated inflammation subsequently modulates glucose metabolic efficiency in future iterations. This bidirectional interaction enables dynamic physiological feedback within the digital twin.

Final estimated biomarker values are determined through the integration of (1) initial screening-day biomarker baselines, (2) stage-specific trend constraints, and (3) physiology-derived process variables generated by the digital twin modules. This integration ensures that longitudinal biomarker trajectories remain physiologically grounded while preserving patient-specific personalization.









\section{Validation of $\htwin$}
\label{sec:validatehepatwin}
In this section, we describe the experiment setups designed to validate the performance of the $\htwin$ model. Fig.~\ref{fig:val1} and ~\ref{fig:val2} demonstrate the high-level overview of the experiment setups. For this work, we utilize the NIDDK NAFLD dataset \cite{tonascia2024nafld}, dividing it into two subsets. One subset serves as the input to $\htwin$, and the second serves as the ground truth for validation. For example, if a patient has 5 longitudinal data points, the first data point is provided as input to $\htwin$, and the model estimates biomarker values for the subsequent 4 visits. These estimated biomarker values are compared with the actual follow-up biomarker values from the NIDDK dataset. The difference between the estimated and ground truth values is computed using Mean Absolute Error (MAE), coefficient of determination (R²), Normalized Root Mean Square Error (NRMSE) and additional statistical metrics.

This validation procedure is repeated across patients based on their available followup information to assess the generalizability and robustness of $\htwin$. Stable error behavior across visits and clinically plausible value ranges provide supporting evidence for model validity.

\begin{figure*}[ht]
\centering
\begin{tabular}{ccc}
\begin{minipage}{0.31\textwidth}
    \centering
    \includegraphics[width=\linewidth]{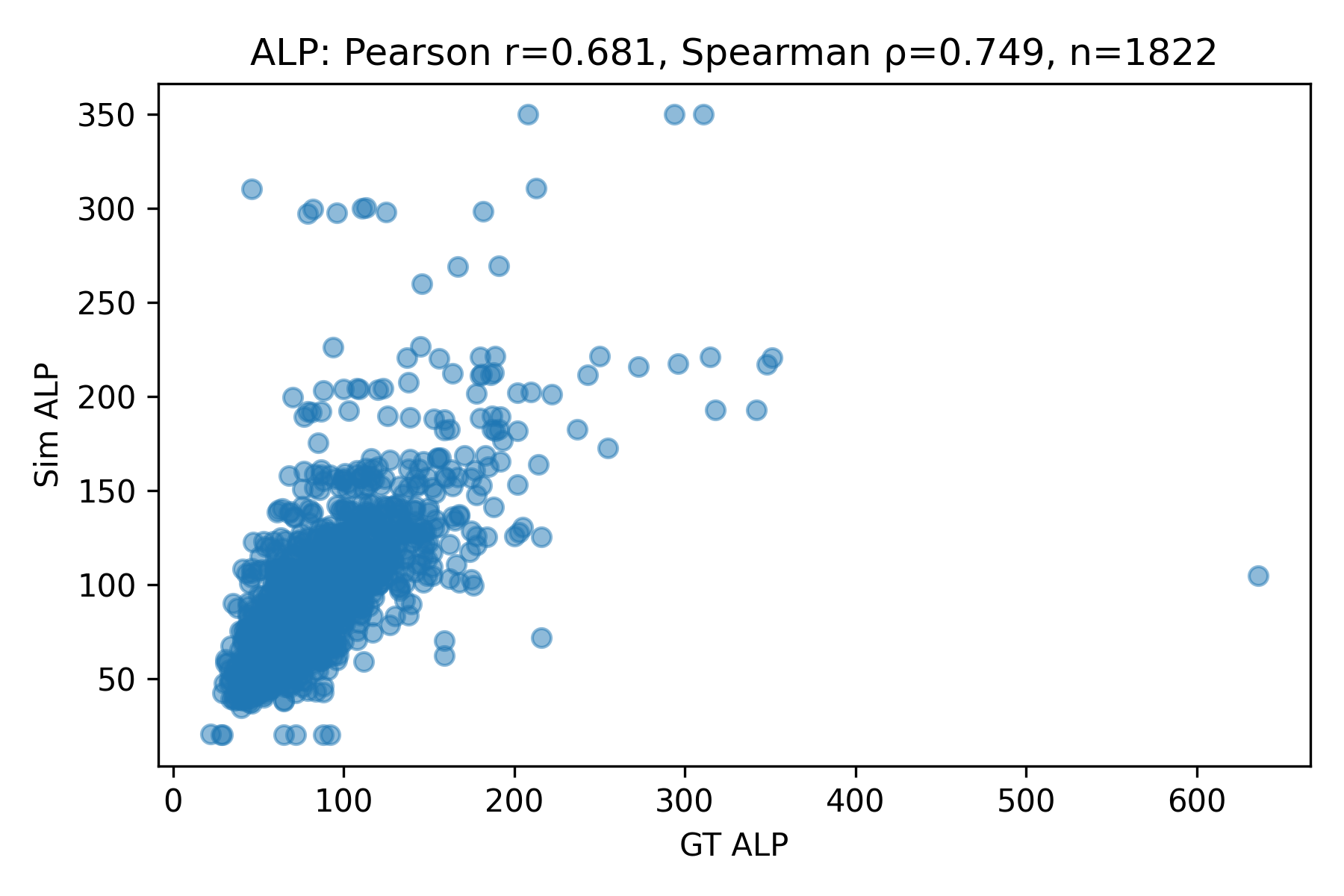}
\end{minipage} &
\begin{minipage}{0.31\textwidth}
    \centering
    \includegraphics[width=\linewidth]{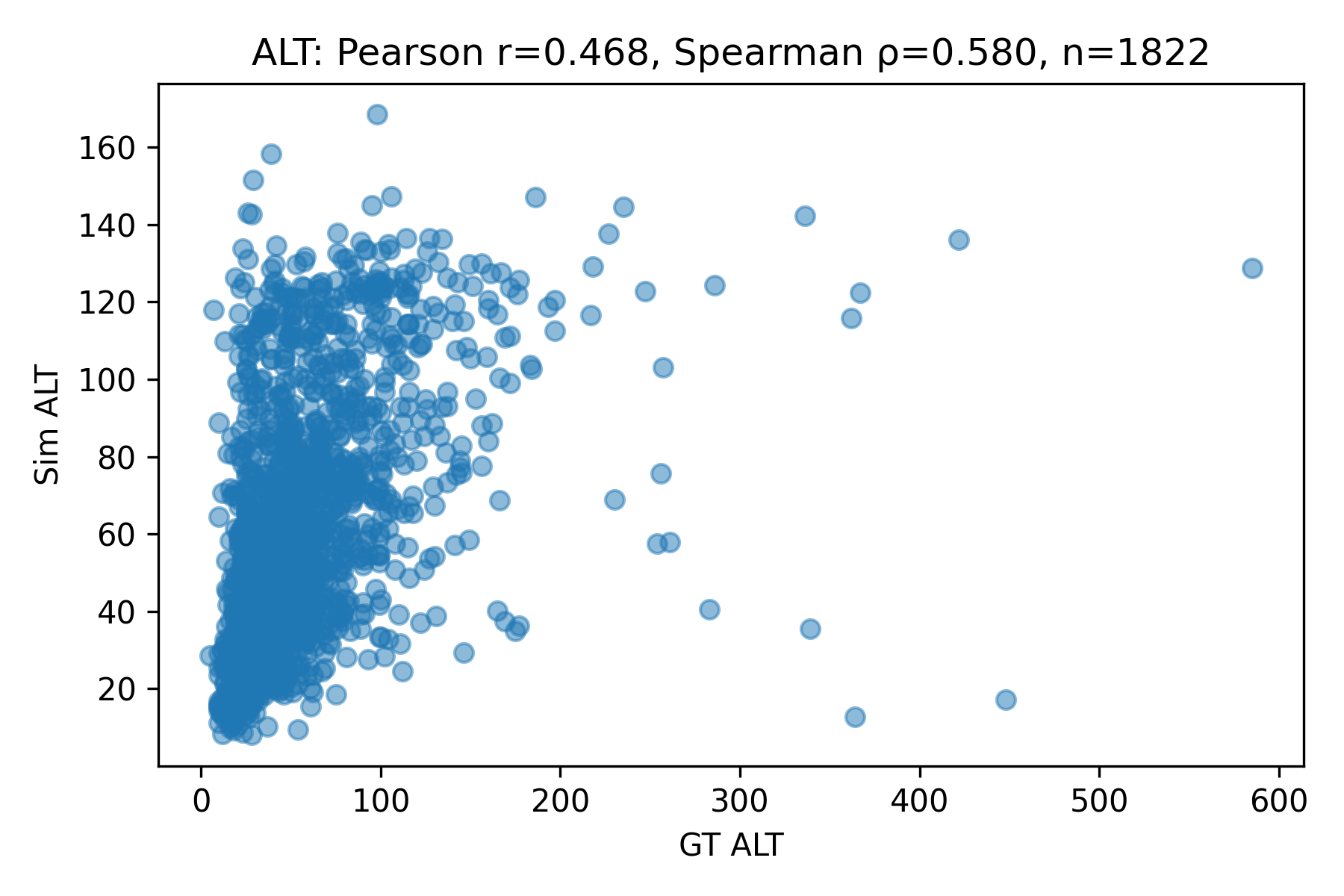}
\end{minipage} &
\begin{minipage}{0.31\textwidth}
    \centering
    \includegraphics[width=\linewidth]{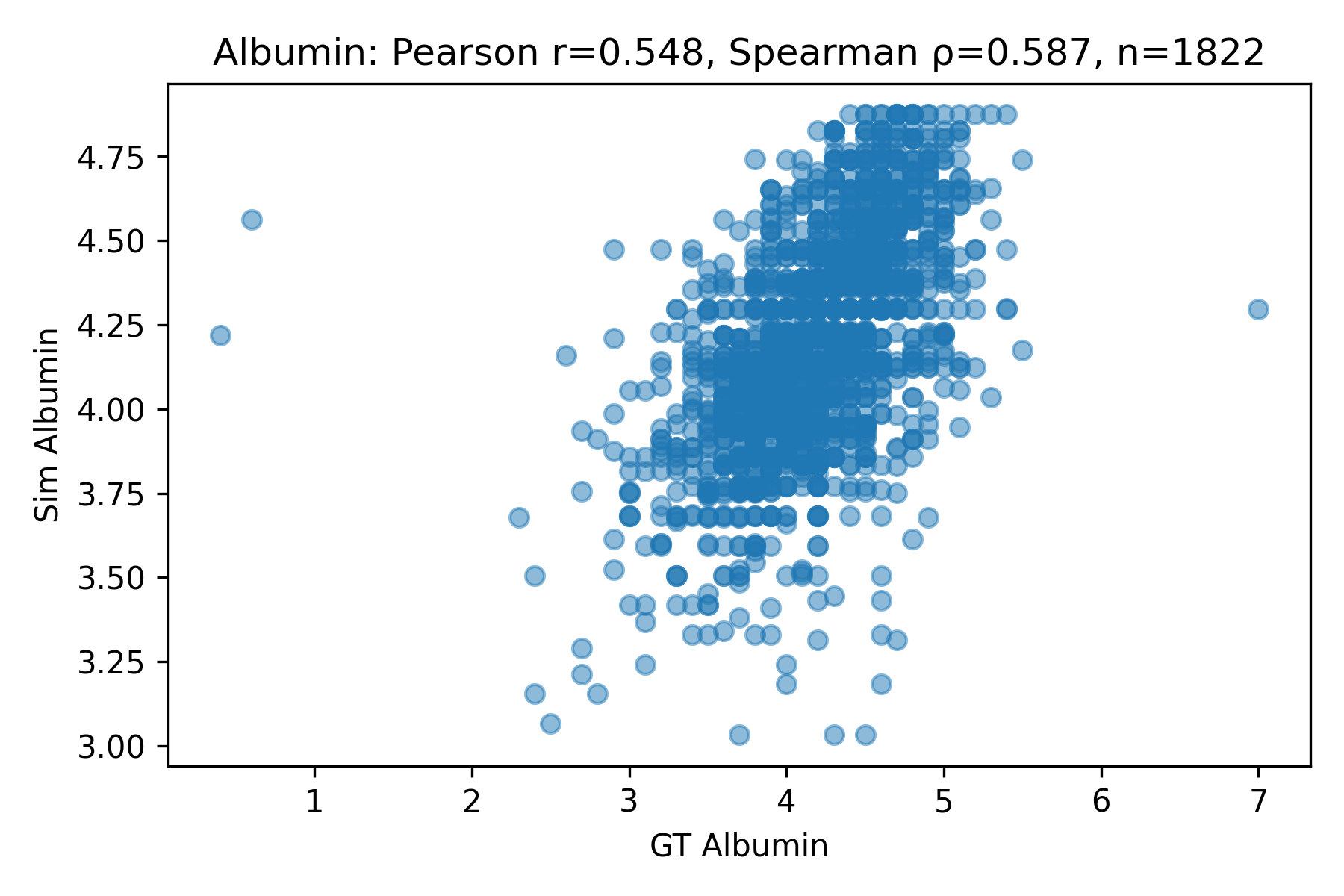}
\end{minipage} \\
(a) ALP & (b) ALT & (c) Albumin \\
\end{tabular}

\vspace{3mm}

\begin{tabular}{cc}
\begin{minipage}{0.32\textwidth}
    \centering
    \includegraphics[width=\linewidth]{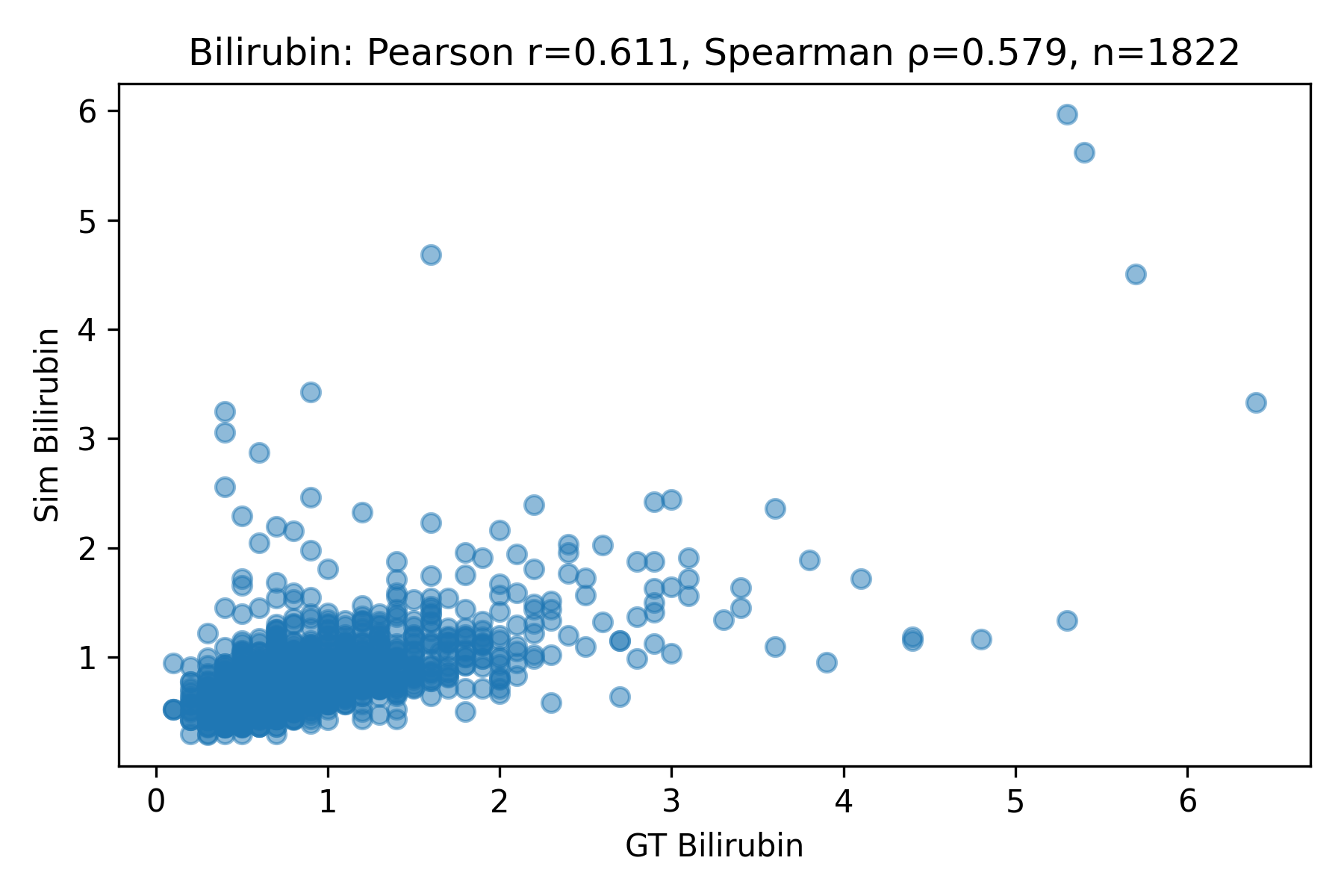}
\end{minipage} &
\begin{minipage}{0.32\textwidth}
    \centering
    \includegraphics[width=\linewidth]{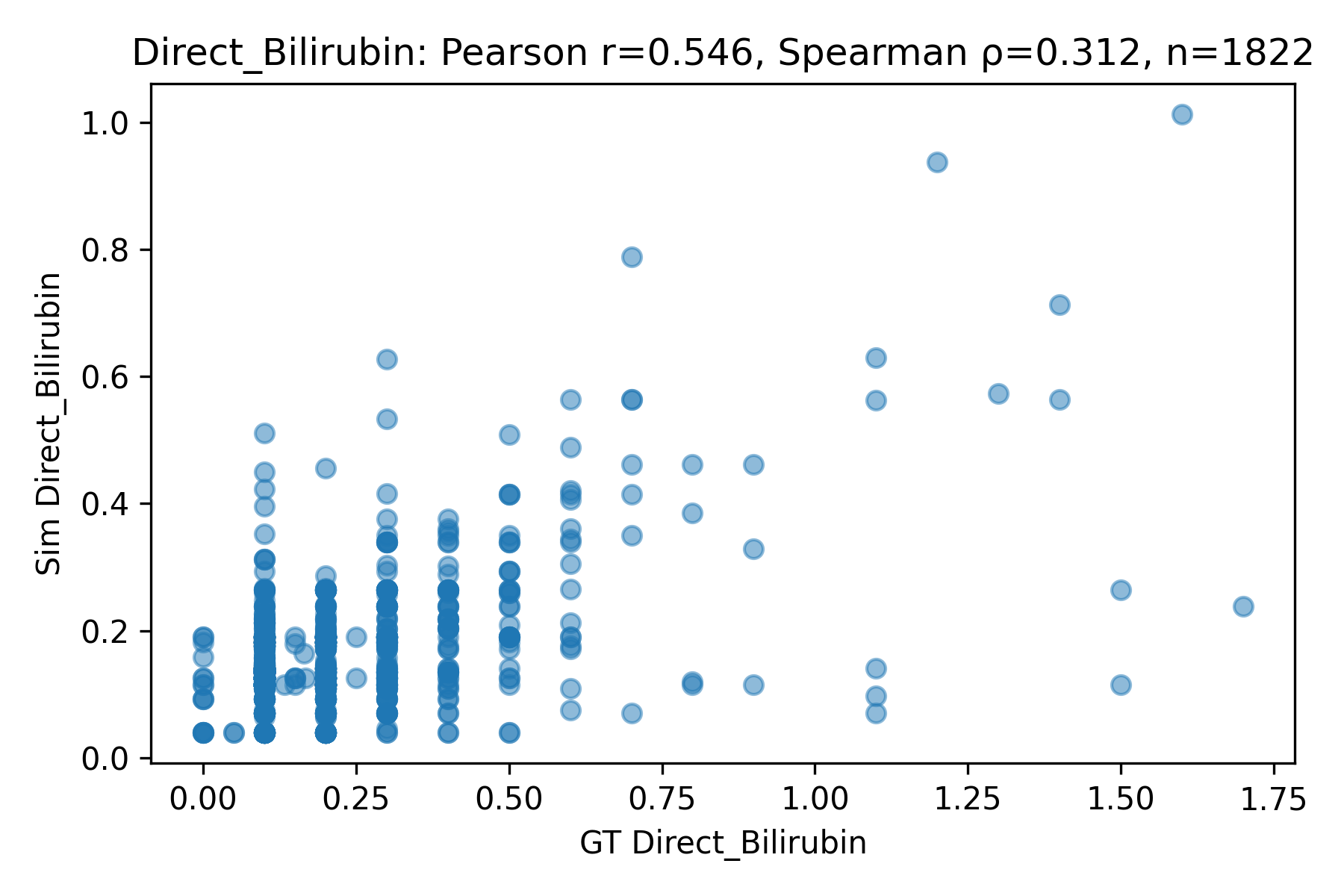}
\end{minipage} \\
(d) Total Bilirubin & (e) Direct Bilirubin \\
\end{tabular}

\caption{Scatter plots showing correlation between $\htwin$-estimated and ground-truth biomarker values across follow-up visits: (a) ALP, (b) ALT, (c) Albumin, (d) Total Bilirubin, and (e) Direct Bilirubin.}
\label{fig:biomarker_scatter}
\end{figure*}

\begin{figure}
\begin{center}
\begin{tabular}{cc}
\begin{minipage}{.5\columnwidth}
    \includegraphics[width=\columnwidth]{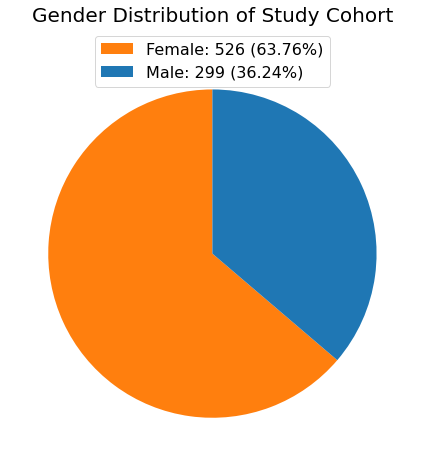}
\end{minipage} &
\begin{minipage}{.5\columnwidth}
    \includegraphics[width=\columnwidth]{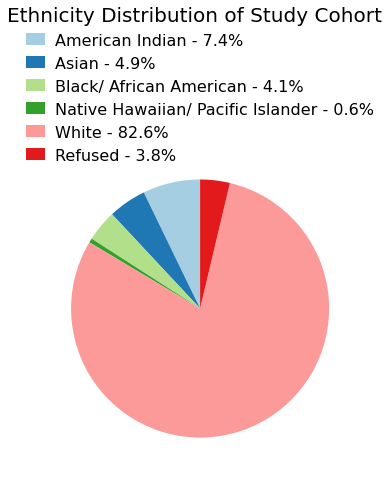}
\end{minipage} 

\\
(a) & (b) 
\end{tabular}
\end{center}
\caption{Distribution of a) Gender, and b) Ethnicity in Study Cohort}
\label{fig:datadist}
\end{figure}

\subsection{NIDDK Dataset and Cohort Selection}

The NIDDK NAFLD dataset contains extensive longitudinal clinical information for each participant. However, since the objective of this work is to analyze the feasibility of NASH detection once NAFLD has been diagnosed using longitudinal lifestyle data, we focus only on the features containing dietary intake, physical activity, laboratory test results, central histology, and registration information.

\begin{table}[t]
\centering
\caption{Selected input features for the $\htwin$ model}
\label{tab:htwin_features}
\begin{tabular}{lll}
\toprule
\textbf{Demographics} & \textbf{Lifestyle} & \textbf{Biomarkers} \\
\midrule
ID & Activity Level & Total Bilirubin \\
Visit & Activity hrs/day & Direct Bilirubin \\
Age & Sitting hrs/day & Albumin \\
Gender & Daily Calories & AST \\
Weight & Protein Intake & ALT \\
Height & Carbohydrate Intake & ALP \\
 & Fat Intake & \\
\bottomrule
\end{tabular}
\end{table}

Table~\ref{tab:htwin_features} lists the final set of selected features for $\htwin$ model. Unique patient identifiers and visit codes are used to align patient information across different features. The NIDDK dataset initially contains 1410 patients.  Since our objective involves forecasting and longitudinal modeling, we retain only patients with at least two visits. This results in 825 participants with minimum two follow-up samples. Patients with only one visit are excluded. Our feature space remains minimal, consisting of basic demographic and lifestyle information, and core liver biomarkers (ALT, ALP, bilirubin, albumin), which are commonly used in early-stage NAFLD/NASH detection and can be collected through minimally invasive procedures. Fig.~\ref{fig:datadist} shows the gender and ethnicity distribution in the study cohort.

\begin{table}[ht]
\centering
\caption{Performance metrics for $\htwin$ predictions using 2 follow-up visits}
\label{tab:metrics_visits_2}
\begin{tabular}{lcccccc}
\toprule
Biomarker & $n$ & MAE & NRMSE$_{range}$ (\%) &  $R^2$ & MAPE (\%) \\
\midrule
AST               & 199 & 14.91 & 9.61  & 0.13 & 36.90 \\
ALT               & 199 & 21.01 & 15.58 & 0.30 & 45.45 \\
ALP               & 199 & 14.81 & 9.96   & 0.52 & 19.10 \\
Albumin           & 199 & 0.29  & 9.32  & 0.22 & 9.70  \\
Bilirubin         & 199 & 0.25  & 10.86 & 0.44 & 37.48 \\
Direct Bili.  & 199 & 0.07  & 7.15   & 0.60 & 32.01 \\
\bottomrule
\end{tabular}
\end{table}

\begin{table}[ht]
\centering
\caption{Performance metrics for $\htwin$ predictions using 3 follow-up visits}
\label{tab:metrics_visits_3}
\begin{tabular}{lccccc}
\toprule
Biomarker & $n$ & MAE & NRMSE$_{range}$ (\%) &  $R^2$ & MAPE (\%) \\
\midrule
AST               & 421 & 17.90 & 11.56  & 0.00 & 44.49 \\
ALT               & 421 & 28.95 & 13.31  & 0.08 & 69.50 \\
ALP               & 421 & 18.96 & 9.70   & 0.41 & 23.67 \\
Albumin           & 421 & 0.29  & 13.10 & 0.26 & 7.34  \\
Bilirubin         & 421 & 0.27  & 9.24   & 0.38 & 33.55 \\
Direct Bili.  & 421 & 0.07  & 8.61   & 0.36 & 32.11 \\
\bottomrule
\end{tabular}
\end{table}

\begin{table}[ht]
\centering
\caption{Performance metrics for $\htwin$ predictions using 4 follow-up visits}
\label{tab:metrics_visits_4}
\begin{tabular}{lccccc}
\toprule
Biomarker & $n$ & MAE & NRMSE$_{range}$ (\%) & $R^2$ & MAPE (\%) \\
\midrule
AST               & 430 & 19.34 & 9.60 & -0.01 & 46.47 \\
ALT               & 430 & 24.30 & 11.48 & 0.14 & 57.22 \\
ALP               & 430 & 19.24 & 9.57   & 0.37 & 23.24 \\
Albumin           & 430 & 0.29  & 8.97   & 0.28 & 7.14  \\
Bilirubin         & 430 & 0.28  & 10.52 & 0.19 & 36.89 \\
Direct Bili.  & 430 & 0.07  & 12.53 & 0.17 & 38.23 \\
\bottomrule
\end{tabular}
\end{table}

\begin{table}[ht]
\centering
\caption{Performance metrics for $\htwin$ predictions using 5 follow-up visits}
\label{tab:metrics_visits_5}
\begin{tabular}{lccccc}
\toprule
Biomarker & $n$ & MAE & NRMSE$_{range}$ (\%) &  $R^2$ & MAPE (\%) \\
\midrule
AST               & 772 & 17.38 & 12.08 & -0.15 & 42.97 \\
ALT               & 772 & 22.24 & 6.82   & 0.20 & 48.33 \\
ALP               & 772 & 19.79 & 5.68   & 0.18 & 24.60 \\
Albumin           & 772 & 0.29  & 7.81    & 0.31 & 8.19  \\
Total Bili.         & 772 & 0.28  & 7.48  & 0.40 & 36.62 \\
Direct Bili.  & 772 & 0.08  & 8.86   & 0.07 & 41.59 \\
\bottomrule
\end{tabular}
\end{table}

\begin{figure*}
\centering

\begin{tabular}{ccc}
\begin{minipage}{0.31\textwidth}
    \centering
    \includegraphics[width=\linewidth]{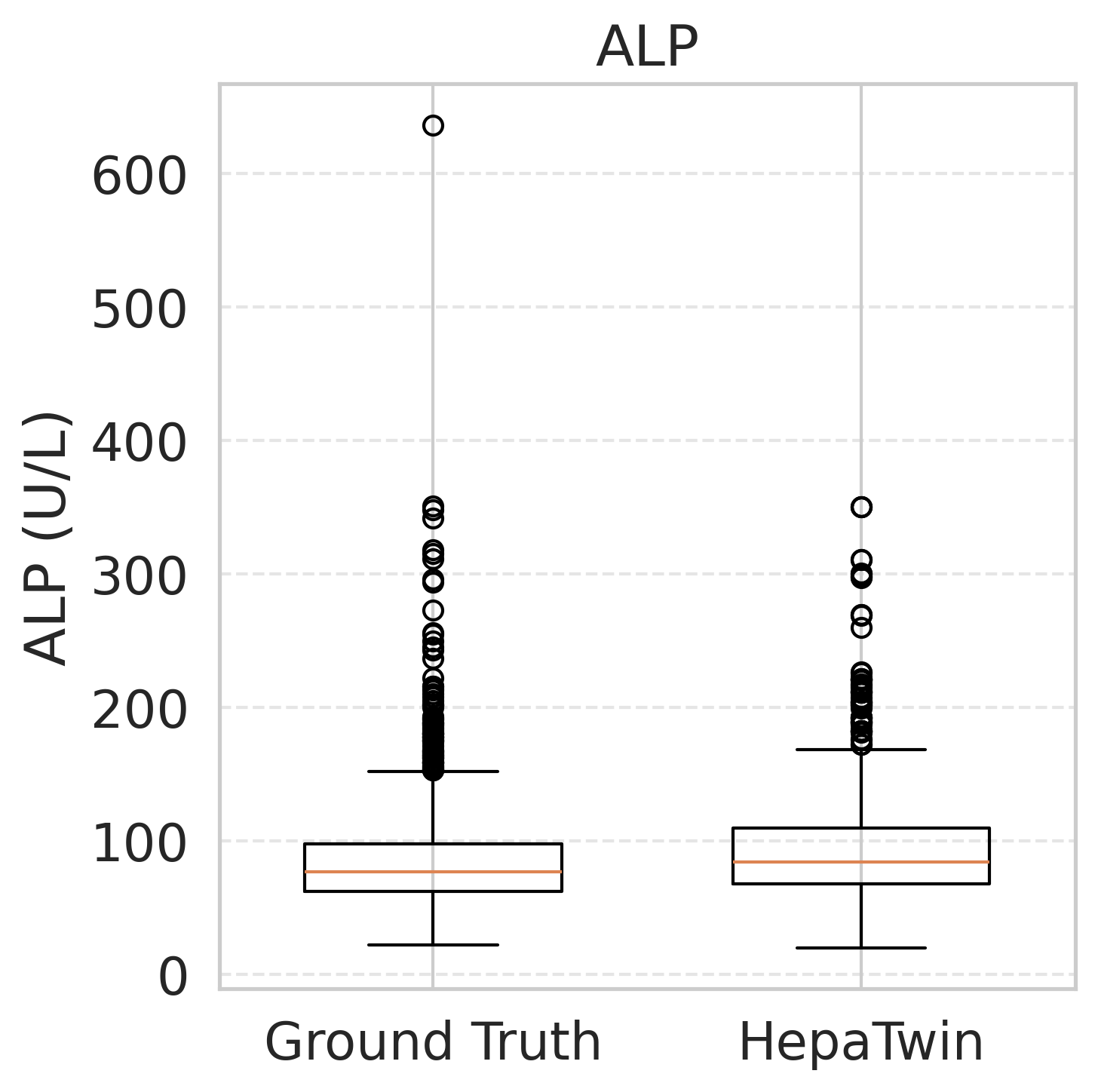}
\end{minipage} &
\begin{minipage}{0.31\textwidth}
    \centering
    \includegraphics[width=\linewidth]{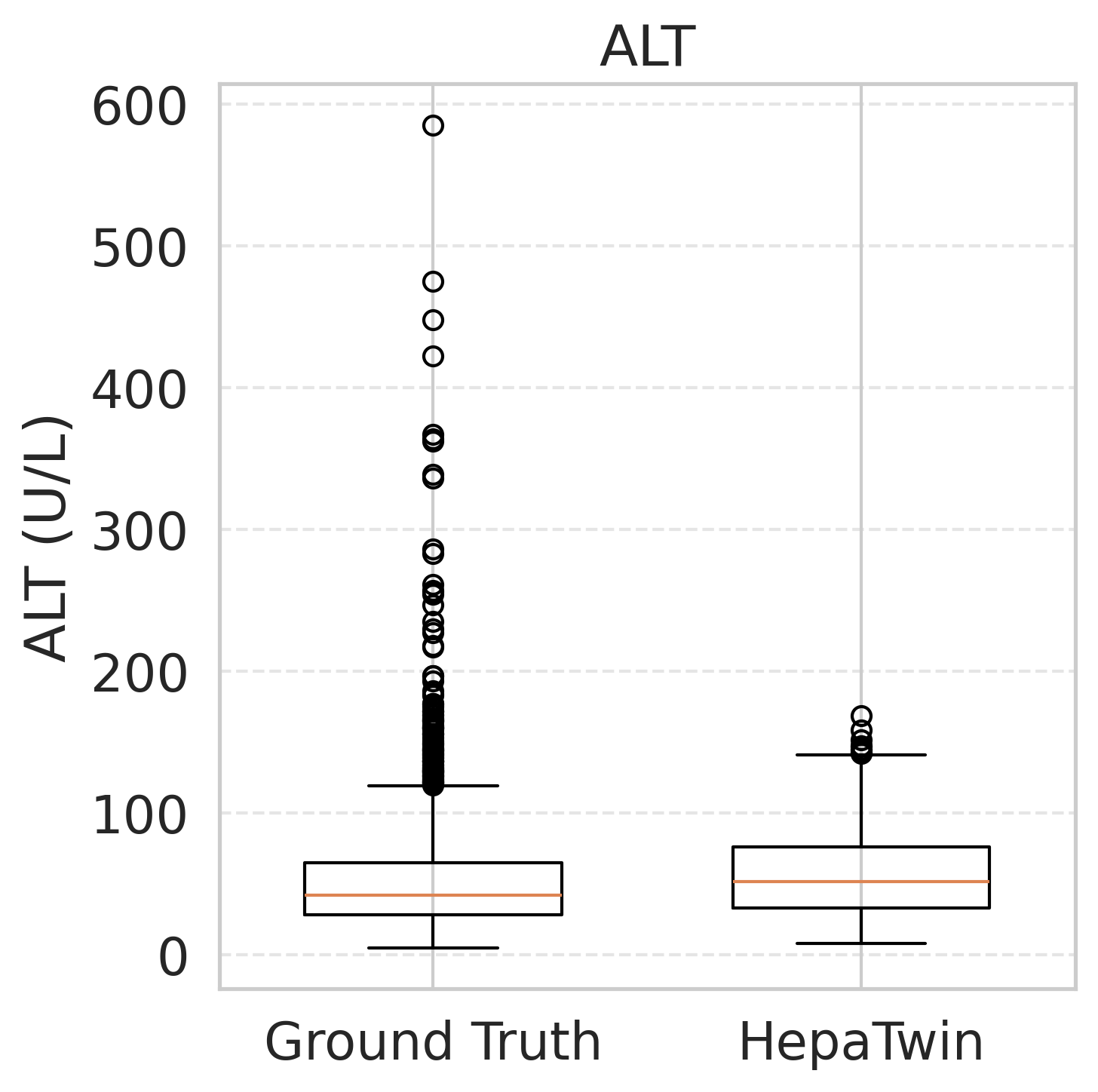}
\end{minipage} &
\begin{minipage}{0.31\textwidth}
    \centering
    \includegraphics[width=\linewidth]{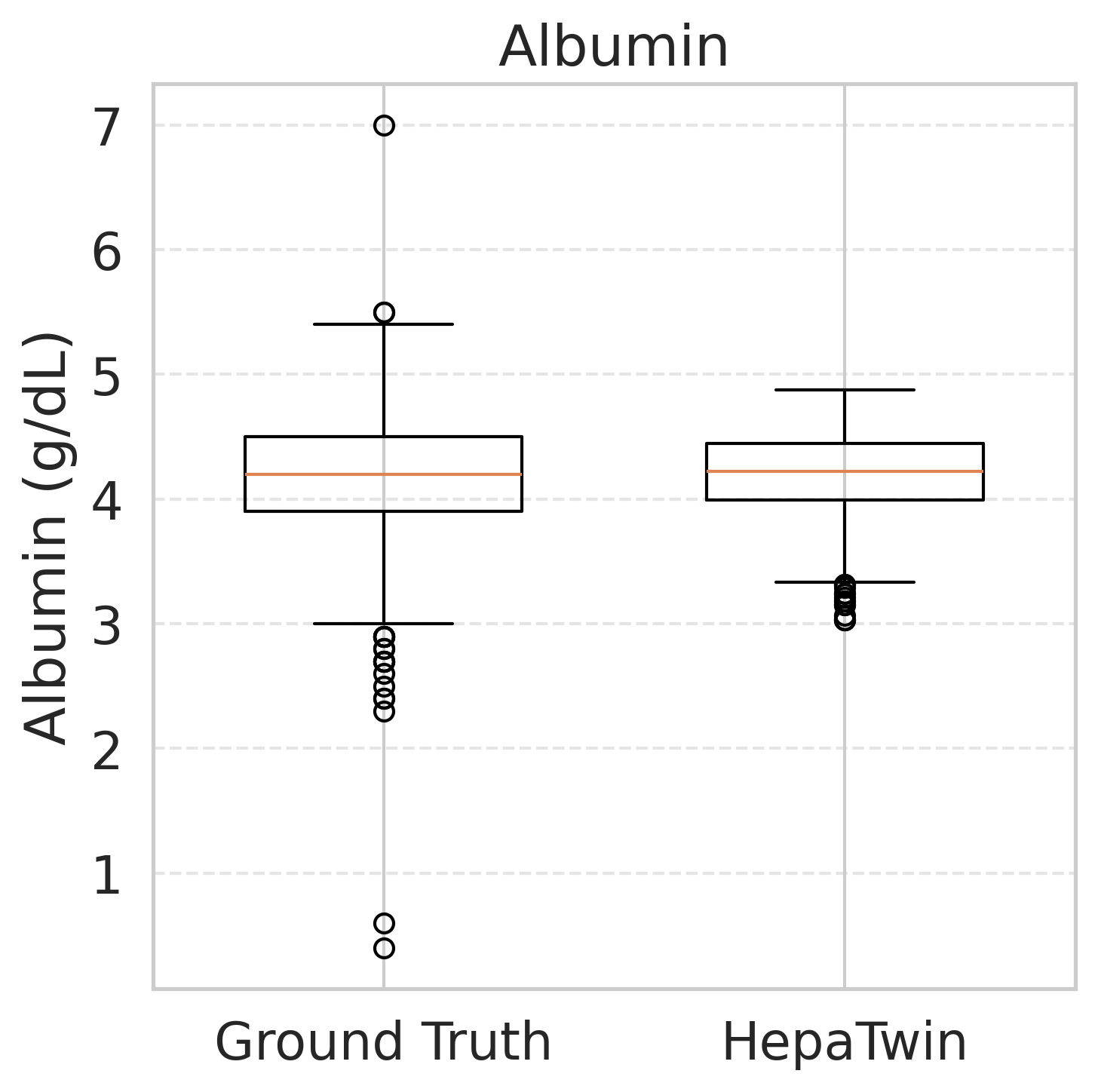}
\end{minipage} \\
(a) ALP & (b) ALT & (c) Albumin \\
\end{tabular}

\vspace{3mm}

\begin{tabular}{cc}
\begin{minipage}{0.31\textwidth}
    \centering
    \includegraphics[width=\linewidth]{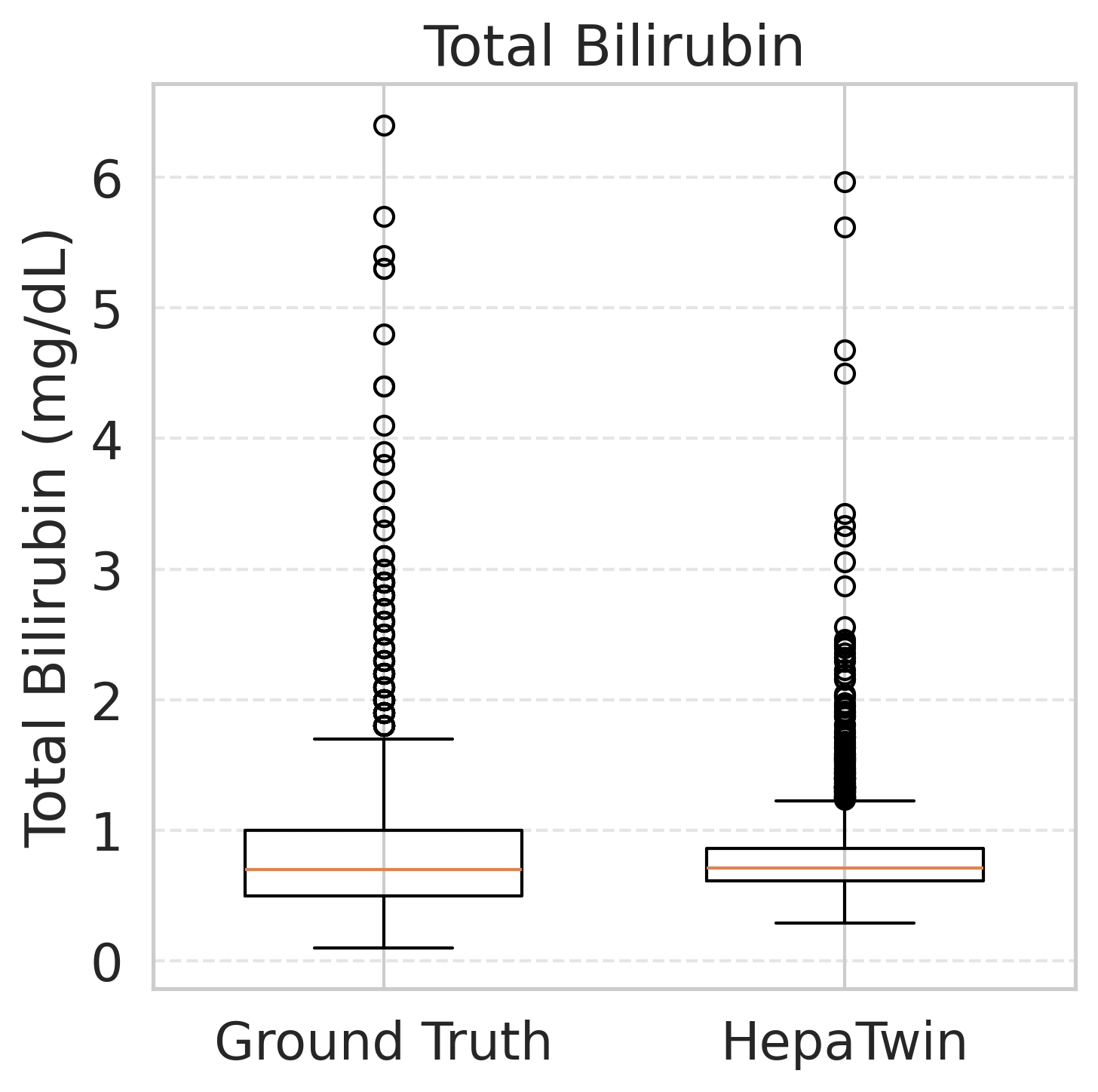}
\end{minipage} &
\begin{minipage}{0.31\textwidth}
    \centering
    \includegraphics[width=\linewidth]{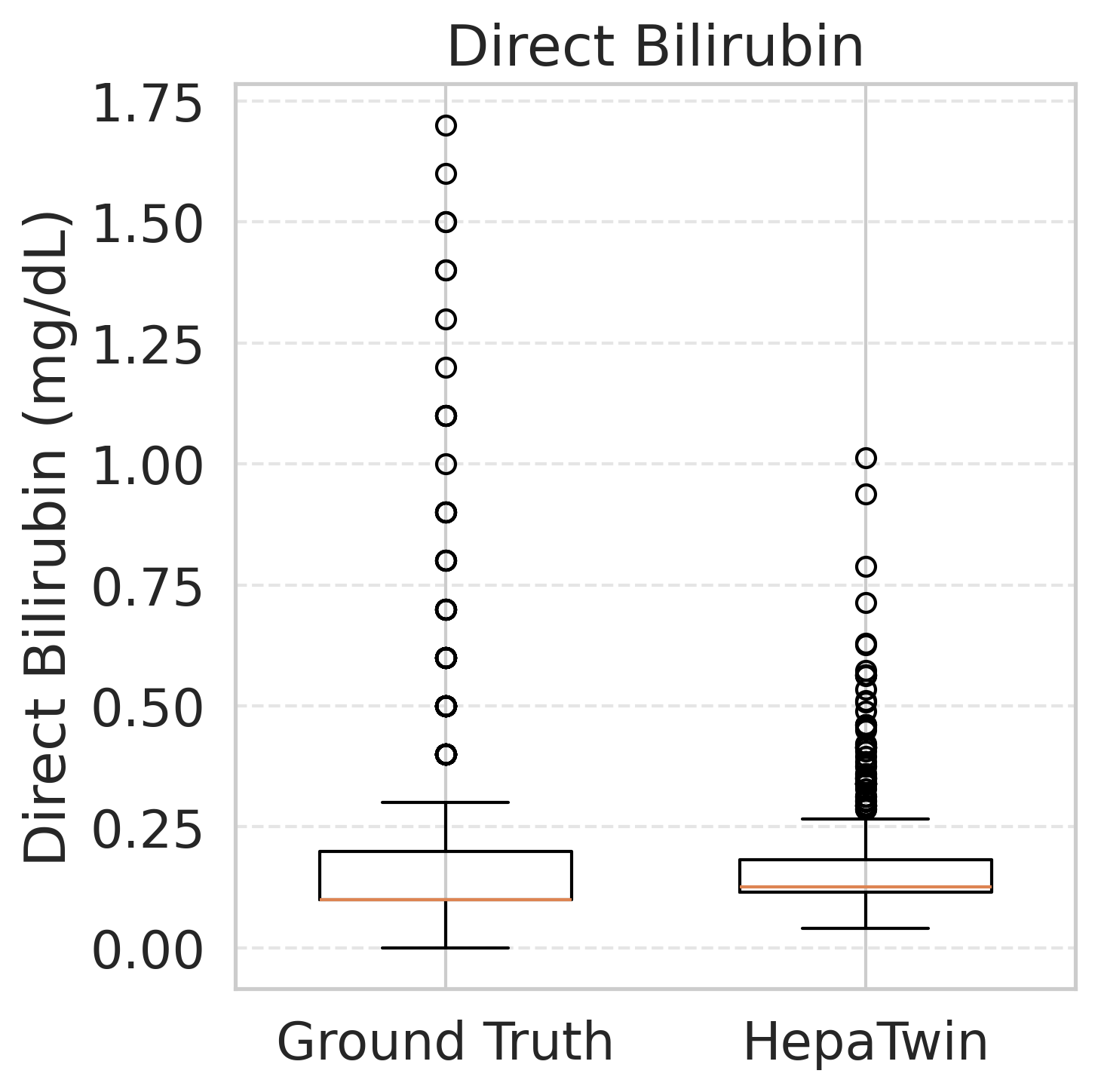}
\end{minipage} \\
(d) Total Bilirubin & (e) Direct Bilirubin \\
\end{tabular}

\caption{Boxplot distributions of $\htwin$-estimated versus ground-truth biomarker values across follow-up visits: (a) ALP, (b) ALT, (c) Albumin, (d) Total Bilirubin, and (e) Direct Bilirubin.}
\label{fig:biomarker_boxplots}
\end{figure*}

\subsection{Validation Strategy I: $\htwin$-Estimated Biomarker Performance across Ground Truth Clinical Biomarker Values}

In the first experiment, we evaluate how closely the $\htwin$ model estimates longitudinal biomarker trajectories to the original clinical real-world data. The real-world NIDDK dataset provides screening-day biomarker values along with full lifestyle information (diet, physical activity, age, gender, height, weight). These variables are used as input to $\htwin$, which executes its physiological modules and generates estimated biomarker values for future visits.

The estimated biomarker values are compared with the NIDDK ground truth follow-up values. Performance is evaluated per biomarker and per number of follow-up visits using standard metrics such as Mean Absolute Error (MAE), R² score, Pearson correlation, Spearman correlation, side-by-side boxplot comparisons of estimated vs ground truth biomarker values.
Tables~\ref{tab:metrics_visits_2}--\ref{tab:metrics_visits_5}
 summarize per-biomarker estimation performance across patients and visits. The results demonstrate that $\htwin$-generated biomarker values fall within clinically reasonable ranges and preserve longitudinal trends observed in the ground truth dataset. Albumin consistently exhibits stable error behavior across increasing visit horizons (${R^2}$ = 0.22–0.31), while transaminases (AST/ALT) show greater variability at extended horizons, likely due to inflammation-driven nonlinear fluctuations not fully captured by the current physiological parameterization. Each visit in the dataset is 48 weeks apart, so forecasting across five visits in $\htwin$ model corresponds to an effective simulation horizon of nearly five years. One possible explanation for the variability in AST estimation is that AST is not exclusively liver-specific; it is also released from cardiac and skeletal muscle tissue, introducing additional physiological variability not explicitly modeled in the current framework. Fig.~\ref{fig:biomarker_scatter} shows the corresponding Pearson and Spearman correlation factors of each biomarker, and Fig.~\ref{fig:biomarker_boxplots} shows side-by-side comparison of ground truth and $\htwin$-estimated biomarker values. Notably, while prior TwinScan (a partial digital twin) results report single-time-point bilirubin estimation errors ranging from 0.27–0.82 across cirrhosis stages \cite{milatwinscan}, $\htwin$ achieves a \textit{\textbf{mean absolute error of 0.28 (total) and 0.08 (direct) even for five-step longitudinal predictions spanning approximately five years, highlighting the robustness of the proposed $\htwin$ model and demonstrating its capability to produce physiologically consistent multi-year biomarker forecasts.}}

\subsection{Validation Strategy II: NASH Detection Using $\htwin$-Estimated Biomarkers}

In the second experiment, we assess whether the estimated biomarker trajectories generated by $\htwin$ retain sufficient clinical signal for NASH detection. First, we establish a baseline by using ground truth biomarker values from the NIDDK dataset to detect NASH status. Using time-series ground truth biomarker values, the detection model achieves a performance of Accuracy: 0.76, Precision: 0.78, Recall: 0.71, F1-score: 0.75. These results are consistent with existing statistical approaches on the same dataset, where models trained using biopsy-day biomarker values achieve approximately 75\% accuracy \cite{docherty2021development}.

Next, we replace ground truth biomarker trajectories with $\htwin$-estimated biomarker trajectories and repeat the NASH detection task. Using the estimated time-series data, the detection accuracy is 65\%. Most existing models trained on the same dataset estimate NASH status using biomarker measurements collected at or near the biopsy visit, where the pathological state of the liver is already reflected in the laboratory values. In contrast, $\htwin$ achieves this \textit{\textbf{65\% NASH detection accuracy}} using \textit{\textbf{baseline lifestyle and demographic information to generate longitudinal biomarker trajectories}} that extend up to \textit{\textbf{five years into the future}}. Achieving 65\% detection accuracy under this long-horizon prediction setting indicates that the physiology-driven digital twin preserves clinically meaningful disease signals over extended time intervals, demonstrating the practical value of $\htwin$ for early risk assessment and long-term liver health monitoring. These findings suggest that even approximate biomarker trajectories derived from $\htwin$ preserve sufficient disease-related signal to support downstream NASH classification.


\subsection{Validation Experiments Takeaway}

Across both experiments, $\htwin$ demonstrates the ability to:
\begin{itemize}
    \item generate longitudinal biomarker estimates that demonstrate stable error behavior and clinically plausible value ranges relative to ground truth values;

    \item preserve clinically relevant disease progression patterns;

\item support downstream NASH detection using simulated biomarker trajectories; and
\item generate physiologically consistent biomarker trajectories extending up to five years into the future using only baseline lifestyle and demographic inputs.
\end{itemize}
These results validate the feasibility of using physiology-informed digital twin modeling for longitudinal liver disease simulation and progression analysis.








\section{Conclusion}
\label{sec:concl}

We presented $\htwin$, a physiology-informed digital twin of the human liver that simulates metabolic, detoxification, and biosynthetic processes to estimate clinically observable biomarker trajectories across disease stages. By integrating carbohydrate metabolism, lipid handling, protein metabolism, bilirubin conjugation, and bile production within a unified systems-level model, $\htwin$ enables mechanistic estimation of biomarkers through evolving internal variables such as hepatocyte functional capacity, inflammation index, cholestasis index, and hepatic fat dynamics. The introduction of a stage-transition-driven progress-factor calibration mechanism ensures that simulated outputs remain consistent with population-level disease progression patterns while preserving physiological interpretability.

Experimental validation using the NIDDK NAFLD dataset demonstrates that $\htwin$ generates longitudinal biomarker trajectories within clinically acceptable ranges and supports forecasting over multi-year horizons. Furthermore, simulated biomarker trajectories retain sufficient clinical signal for downstream NASH detection, achieving competitive performance relative to models based on biopsy-day measurements. Beyond biomarker estimation, $\htwin$ establishes a foundation for closed-loop decision support by enabling integration with machine learning models to provide interpretable insights into the impact of lifestyle factors on liver health. Overall, this work highlights the potential of combining physiology-based modeling with data-driven calibration to enable scalable, non-invasive, and personalized liver health monitoring,

Future work will focus on extending the digital twin to incorporate additional inter-organ interactions and adaptive feedback mechanisms to further improve physiological realism and personalized lifestyle guidance.

\section*{Data Availability}
No new data was collected in support of this research. The data from the NAFLD Adult Database used here were supplied by the NIDDK Central Repository\cite{tonascia2024nafld}.

\section*{References}
\bibliographystyle{IEEEtran}
\bibliography{IEEEreference}

\begin{thebibliography}{10}
\providecommand{\url}[1]{#1}
\csname url@samestyle\endcsname
\providecommand{\newblock}{\relax}
\providecommand{\bibinfo}[2]{#2}
\providecommand{\BIBentrySTDinterwordspacing}{\spaceskip=0pt\relax}
\providecommand{\BIBentryALTinterwordstretchfactor}{4}
\providecommand{\BIBentryALTinterwordspacing}{\spaceskip=\fontdimen2\font plus
\BIBentryALTinterwordstretchfactor\fontdimen3\font minus \fontdimen4\font\relax}
\providecommand{\BIBforeignlanguage}[2]{{%
\expandafter\ifx\csname l@#1\endcsname\relax
\typeout{** WARNING: IEEEtran.bst: No hyphenation pattern has been}%
\typeout{** loaded for the language `#1'. Using the pattern for}%
\typeout{** the default language instead.}%
\else
\language=\csname l@#1\endcsname
\fi
#2}}
\providecommand{\BIBdecl}{\relax}
\BIBdecl

\bibitem{b1}
\BIBentryALTinterwordspacing
Y.~B. Liu and M.-K. Chen, ``Epidemiology of liver cirrhosis and associated complications: Current knowledge and future directions,'' \emph{World Journal of Gastroenterology}, vol.~28, no.~41, pp. 5910--5930, Nov 2022, pMCID: PMC9669831. [Online]. Available: \url{https://www.ncbi.nlm.nih.gov/pmc/articles/PMC9669831/}
\BIBentrySTDinterwordspacing

\bibitem{b2}
S.~Scaglione, S.~Kliethermes, G.~Cao, D.~Shoham, R.~Durazo, A.~Luke, and M.~L. Volk, ``The epidemiology of cirrhosis in the united states: A population-based study,'' \emph{Journal of Clinical Gastroenterology}, vol.~49, no.~8, pp. 690--696, September 2015.

\bibitem{milamasc}
S.~A. Mila, B.~B.~Y. Ravi, M.~R. Kabir, and S.~Ray, ``Masc: wearable design for infectious disease detection through machine learning,'' \emph{IEEE Access}, vol.~13, pp. 24\,108--24\,123, 2025.

\bibitem{milatwinscan}
S.~A. Mila and S.~Ray, ``Twin-scan: Liver biomarker estimation using machine learning and digital twin simulation,'' in \emph{2025 IEEE EMBS International Conference on Biomedical and Health Informatics (BHI)}, 2025, pp. 1--7.

\bibitem{kandalgaonkar2024digestive}
M.~R. Kandalgaonkar, V.~Kumar, and M.~Vijay-Kumar, ``Digestive dynamics: Unveiling interplay between the gut microbiota and the liver in macronutrient metabolism and hepatic metabolic health,'' \emph{Physiological Reports}, vol.~12, no.~12, p. e16114, 2024.

\bibitem{solomando2022microplastic}
A.~Solomando, A.~Cohen-S{\'a}nchez, A.~Box, I.~Montero, S.~Pinya, and A.~Sureda, ``Microplastic presence in the pelagic fish, seriola dumerili, from balearic islands (western mediterranean), and assessment of oxidative stress and detoxification biomarkers in liver,'' \emph{Environmental research}, vol. 212, p. 113369, 2022.

\bibitem{lopez2022contribution}
L.~L{\'o}pez-Bermudo, A.~Luque-Sierra, D.~Maya-Miles, R.~Gallego-Dur{\'a}n, J.~Ampuero, M.~Romero-G{\'o}mez, G.~Bern{\'a}, and F.~Mart{\'\i}n, ``Contribution of liver and pancreatic islet crosstalk to $\beta$-cell function/dysfunction in the presence of fatty liver,'' \emph{Frontiers in endocrinology}, vol.~13, p. 892672, 2022.

\bibitem{guo2023metabolic}
S.~Guo, Y.~Feng, X.~Zhu, X.~Zhang, H.~Wang, R.~Wang, Q.~Zhang, Y.~Li, Y.~Ren, X.~Gao \emph{et~al.}, ``Metabolic crosstalk between skeletal muscle cells and liver through irf4-fstl1 in nonalcoholic steatohepatitis,'' \emph{Nature Communications}, vol.~14, no.~1, p. 6047, 2023.

\bibitem{gilani2024adipose}
A.~Gilani, L.~Stoll, E.~A. Homan, and J.~C. Lo, ``Adipose signals regulating distal organ health and disease,'' \emph{Diabetes}, vol.~73, no.~2, pp. 169--177, 2024.

\bibitem{IQWiG2021}
{Institute for Quality and Efficiency in Health Care (IQWiG)}, ``In brief: How does the gallbladder work?'' Cologne, Germany, 2021, updated April 27, 2021. Available at: https://www.ncbi.nlm.nih.gov/books/NBK279386/.

\bibitem{Jensen2011Glycogen}
J.~Jensen, P.~I. Rustad, A.~J. Kolnes, and Y.-C. Lai, ``The role of skeletal muscle glycogen breakdown for regulation of insulin sensitivity by exercise,'' \emph{Frontiers in Physiology}, vol.~2, p. 112, 2011.

\bibitem{Wasserman2009FourGrams}
D.~H. Wasserman, ``Four grams of glucose,'' \emph{American Journal of Physiology - Endocrinology and Metabolism}, vol. 296, no.~1, pp. E11--E21, 2009.

\bibitem{glucosemetabolism}
\BIBentryALTinterwordspacing
M.~Adeva-Andany, N.~Pérez-Felpete, C.~Fernández-Fernández, C.~Donapetry-García, and C.~Pazos-García, ``Liver glucose metabolism in humans,'' \emph{Bioscience Reports}, vol.~36, no.~6, p. e00416, 11 2016. [Online]. Available: \url{https://doi.org/10.1042/BSR20160385}
\BIBentrySTDinterwordspacing

\bibitem{veldhorst2009gluconeogenesis}
M.~A. Veldhorst, M.~S. Westerterp-Plantenga, and K.~R. Westerterp, ``Gluconeogenesis and energy expenditure after a high-protein, carbohydrate-free diet,'' \emph{The American journal of clinical nutrition}, vol.~90, no.~3, pp. 519--526, 2009.

\bibitem{Sanvictores2025Fasting}
T.~Sanvictores, J.~Casale, and M.~R. Huecker, ``Physiology, fasting,'' In: StatPearls [Internet], Treasure Island (FL), 2025, updated July 24, 2023. Available from: https://www.ncbi.nlm.nih.gov/books/NBK534877/.

\bibitem{BROWN2017412}
\BIBentryALTinterwordspacing
D.~L. Brown, A.~J. {Van Wettere}, and J.~M. Cullen, ``Chapter 8 - hepatobiliary system and exocrine pancreas1,'' in \emph{Pathologic Basis of Veterinary Disease (Sixth Edition)}, sixth edition~ed., J.~F. Zachary, Ed.\hskip 1em plus 0.5em minus 0.4em\relax Mosby, 2017, pp. 412--470.e1. [Online]. Available: \url{https://www.sciencedirect.com/science/article/pii/B9780323357753000084}
\BIBentrySTDinterwordspacing

\bibitem{ramirez2024multifaceted}
M.~M. Ram{\'\i}rez-Mej{\'\i}a, S.~M. Castillo-Casta{\~n}eda, S.~C. Pal, X.~Qi, and N.~M{\'e}ndez-S{\'a}nchez, ``The multifaceted role of bilirubin in liver disease: a literature review,'' \emph{Journal of Clinical and Translational Hepatology}, vol.~12, no.~11, p. 939, 2024.

\bibitem{Kalakonda2022Bilirubin}
\BIBentryALTinterwordspacing
A.~Kalakonda, B.~A. Jenkins, and S.~John, ``Physiology, bilirubin,'' in \emph{StatPearls [Internet]}.\hskip 1em plus 0.5em minus 0.4em\relax Treasure Island (FL): StatPearls Publishing, 2025, [Updated 2022 Sep 12]. [Online]. Available: \url{https://www.ncbi.nlm.nih.gov/books/NBK470290/}
\BIBentrySTDinterwordspacing

\bibitem{macena2023estimates}
M.~L. Macena, A.~E. Silva~Junior, J.~M. Melo, D.~T. Paula, D.~R. Praxedes, and N.~B. Bueno, ``Estimates of resting energy expenditure and total energy expenditure using predictive equations for individuals after bariatric surgery: a systematic review with meta-analysis,'' \emph{Obesity Surgery}, vol.~33, no.~12, pp. 3999--4006, 2023.

\bibitem{volek2024expert}
J.~S. Volek, W.~S. Yancy~Jr, B.~A. Gower, S.~D. Phinney, J.~Slavin, A.~P. Koutnik, M.~Hurn, J.~Spinner, M.~Cucuzzella, and F.~M. Hecht, ``Expert consensus on nutrition and lower-carbohydrate diets: An evidence-and equity-based approach to dietary guidance,'' \emph{Frontiers in Nutrition}, vol.~11, p. 1376098, 2024.

\bibitem{schaefer2009dietary}
E.~J. Schaefer, J.~A. Gleason, and M.~L. Dansinger, ``Dietary fructose and glucose differentially affect lipid and glucose homeostasis,'' \emph{The Journal of nutrition}, vol. 139, no.~6, pp. 1257S--1262S, 2009.

\bibitem{lu2024effects}
D.~Lu, Y.~Liu, M.~Zhao, S.~Yuan, D.~Liu, X.~Wang, Y.~Liu, Y.~Zhang, M.~Li, Y.~L{\"u} \emph{et~al.}, ``Effects of different doses of glucose and fructose on central carbon metabolic pathways and intercellular wireless communication networks in humans,'' \emph{Food Science and Human Wellness}, vol.~13, no.~4, pp. 1906--1916, 2024.

\bibitem{THALACKERMERCER20071734}
\BIBentryALTinterwordspacing
A.~E. Thalacker-Mercer, C.~A. Johnson, K.~E. Yarasheski, N.~S. Carnell, and W.~W. Campbell, ``Nutrient ingestion, protein intake, and sex, but not age, affect the albumin synthesis rate in humans123,'' \emph{The Journal of Nutrition}, vol. 137, no.~7, pp. 1734--1740, 2007. [Online]. Available: \url{https://www.sciencedirect.com/science/article/pii/S0022316622093063}
\BIBentrySTDinterwordspacing

\bibitem{paulusma2022amino}
C.~C. Paulusma, W.~H. Lamers, S.~Broer, and S.~F. van~de Graaf, ``Amino acid metabolism, transport and signalling in the liver revisited,'' \emph{Biochemical pharmacology}, vol. 201, p. 115074, 2022.

\bibitem{ZOU2023275}
\BIBentryALTinterwordspacing
P.~Zou and L.~Wang, ``Dietary pattern and hepatic lipid metabolism,'' \emph{Liver Research}, vol.~7, no.~4, pp. 275--284, 2023. [Online]. Available: \url{https://www.sciencedirect.com/science/article/pii/S2542568423000673}
\BIBentrySTDinterwordspacing

\bibitem{Nguyenliver}
\BIBentryALTinterwordspacing
P.~Nguyen, V.~Leray, M.~Diez, S.~Serisier, J.~L. Bloc’h, B.~Siliart, and H.~Dumon, ``Liver lipid metabolism,'' \emph{Journal of Animal Physiology and Animal Nutrition}, vol.~92, no.~3, pp. 272--283. [Online]. Available: \url{https://onlinelibrary.wiley.com/doi/abs/10.1111/j.1439-0396.2007.00752.x}
\BIBentrySTDinterwordspacing

\bibitem{ahmed2022functional}
M.~Ahmed, ``Functional, diagnostic and therapeutic aspects of bile,'' \emph{Clinical and experimental gastroenterology}, pp. 105--120, 2022.

\bibitem{Hundt2022BileSecretion}
\BIBentryALTinterwordspacing
M.~Hundt, H.~Basit, and S.~John, ``Physiology, bile secretion,'' in \emph{StatPearls [Internet]}.\hskip 1em plus 0.5em minus 0.4em\relax Treasure Island (FL): StatPearls Publishing, 2025, [Updated 2022 Sep 26]. [Online]. Available: \url{https://www.ncbi.nlm.nih.gov/books/NBK470209/}
\BIBentrySTDinterwordspacing

\bibitem{cholestasis}
\BIBentryALTinterwordspacing
N.~Tripathi and I.~Jialal, ``Conjugated hyperbilirubinemia,'' 2025, [Updated 2023 Jul 24]. Treasure Island (FL): StatPearls Publishing. [Online]. Available: \url{https://www.ncbi.nlm.nih.gov/books/NBK562172/}
\BIBentrySTDinterwordspacing

\bibitem{cirrhosis_patient_survival_prediction_878}
E.~Dickson, P.~Grambsch, T.~Fleming, L.~Fisher, and A.~Langworthy, ``{Cirrhosis Patient Survival Prediction},'' UCI Machine Learning Repository, 1989, {DOI}: https://doi.org/10.24432/C5R02G.

\bibitem{tonascia2024nafld}
\BIBentryALTinterwordspacing
J.~Tonascia, ``Nonalcoholic fatty liver disease (nafld) adult database (nafld adult) (version 4),'' 2024, dataset. [Online]. Available: \url{https://doi.org/10.58020/53bk-jk73}
\BIBentrySTDinterwordspacing

\bibitem{docherty2021development}
M.~Docherty, S.~A. Regnier, G.~Capkun, M.-M. Balp, Q.~Ye, N.~Janssens, A.~Tietz, J.~L{\"o}ffler, J.~Cai, M.~C. Pedrosa \emph{et~al.}, ``Development of a novel machine learning model to predict presence of nonalcoholic steatohepatitis,'' \emph{Journal of the American Medical Informatics Association}, vol.~28, no.~6, pp. 1235--1241, 2021.

\end{thebibliography}

\begin{IEEEbiography}[{\includegraphics[width=1in,height=1.25in,clip,keepaspectratio]{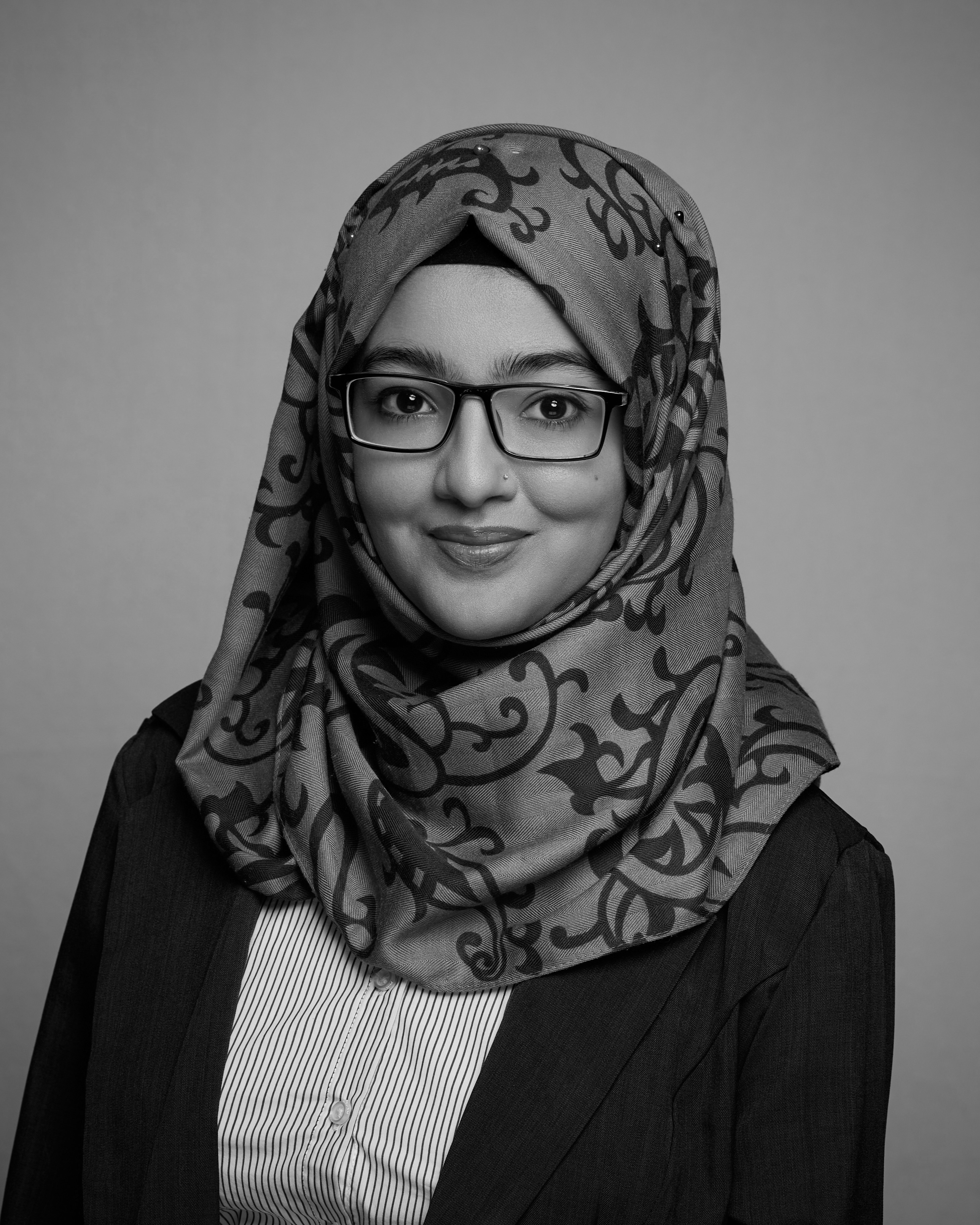}}]{Sumaiya Afroz Mila} received her B.Sc. degree in Computer Science \& engineering from the Military Institue of Science and Technology, Dhaka, Bangladesh and her M.S. degree in Computer Engineering from the University of Florida, USA. She is currently pursuing her Ph.D. degree
in the Department of Electrical and Computer
Engineering, University of Florida. She is also a Research Assistant in the Rising Laboratory, University of Florida. Her research interests include IoT and machine learning applications, digital twin modeling for organ physiology simulation, clinical data analysis, and wearable design.
\end{IEEEbiography}

\begin{IEEEbiography} [{\includegraphics[width=1in,height=1.25in,clip,keepaspectratio]{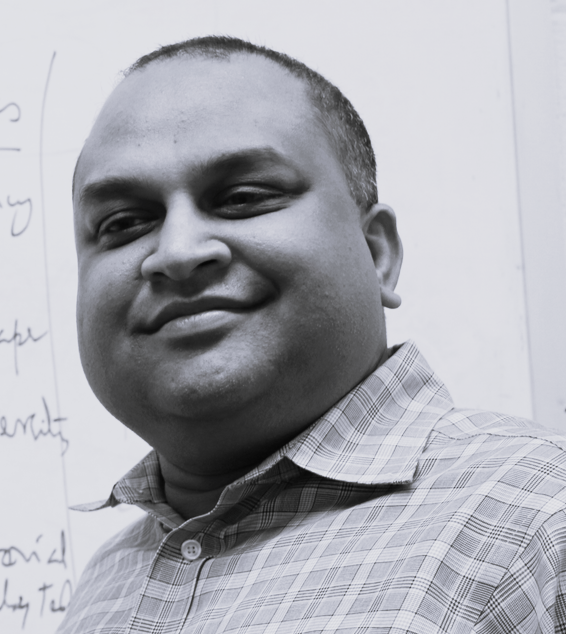}}]{Sandip Ray}  (Senior Member, IEEE) is a Professor at the Department of Electrical and Computer Engineering, University of Florida, Gainesville, FL, where he holds an Endowed IoT Term Professorship.  Before joining the University of Florida, he was a Senior Principal Engineer at NXP Semiconductors, and prior to that, Research Scientist with the Intel Strategic CAD Laboratories.  Dr. Ray’s current research targets correct, dependable, secure, and trustworthy computing through the cooperation of specification, synthesis, architecture, and validation technologies. He is the author of three books and over 100 publications in international journals and conferences.  He has also served as a Technical Program Committee Member of over 50 international conferences, as Program Chair of ACL2 2009, FMCAD 2013, and IFIP IoT 2019, as Guest Editor for IEEE Design and Test, IEEE TMSCS, and ACM TODAES, and as Associate Editor of Springer HaSS and IEEE TMSCS. Dr. Ray has a Ph.D. from the University of Texas at Austin and is a Senior Member of IEEE. 
\end{IEEEbiography}

\end{document}